\documentclass[journal]{IEEEtran}

\usepackage{times}
\usepackage{epsfig}
\usepackage{epstopdf}
\usepackage{graphicx}
\usepackage{amsmath}
\usepackage{amssymb}
\usepackage{color}
\usepackage[english]{babel}

\usepackage{algorithm} %format of the algorithm
\usepackage{algpseudocode} %format of the algorithm
\usepackage{booktabs}
\usepackage{verbatim}
\usepackage{gensymb}
\usepackage{pgffor}

\usepackage{xspace}

\makeatletter
\DeclareRobustCommand\onedot{\futurelet\@let@token\@onedot}
\def\@onedot{\ifx\@let@token.\else.\null\fi\xspace}

\def\eg{{e.g}\onedot} 
\def\ie{{i.e}\onedot} 
 
\def\etc{{etc}\onedot} \def\vs{{vs}\onedot}
 
\def\etal{{et al}\onedot}
\makeatother

\makeatletter
\let\MYcaption\@makecaption
\makeatother

\usepackage[font=footnotesize]{subcaption}

\makeatletter
\let\@makecaption\MYcaption
\makeatother

\makeatletter
\def\zapcolorreset{\let\reset@color\relax\ignorespaces}
\def\colorrows#1{\noalign{\aftergroup\zapcolorreset#1}\ignorespaces}
\makeatother

\begin{document}
\title{Combining Data-driven and Model-driven Methods for Robust Facial Landmark Detection}

\author{Hongwen Zhang,
        Qi Li,
        Zhenan Sun,~\IEEEmembership{Member,~IEEE}, and % <-this % stops a space
        Yunfan Liu
\thanks{This work was supported in part by the National Key Research and Development Program of China under Grant 2016YFB1001000, in part by the National Natural Science Foundation of China under Grant 61573360, Grant 61427811 and Grant 61702513. (Corresponding author: Qi Li, Zhenan Sun.)}
\thanks{H. Zhang, Q. Li, Z. Sun, and Y. Liu are with the Center for Research on Intelligent Perception and Computing,
	National Laboratory of Pattern Recognition, Institute of Automation,
	CAS Center for Excellence in Brain Science and Intelligence Technology,
	Chinese Academy of Sciences,
	100190, Beijing, China.
	E-mail: hongwen.zhang@cripac.ia.ac.cn;
	qli@nlpr.ia.ac.cn;
	znsun@nlpr.ia.ac.cn;
	yunfan.liu@cripac.ia.ac.cn.}% <-this % stops a space
\thanks{H. Zhang and Z. Sun are also with the University of Chinese Academy of Sciences, Beijing, China.}
}

\markboth{IEEE TRANSACTIONS ON INFORMATION FORENSICS AND SECURITY}%
{Shell \MakeLowercase{\textit{et al.}}: Bare Demo of IEEEtran.cls for IEEE Journals}

\maketitle

\begin{abstract}
Facial landmark detection is an important yet challenging task for real-world computer vision applications. This paper proposes an effective and robust approach for facial landmark detection by combining data- and model-driven methods. Firstly, a Fully Convolutional Network (FCN) is trained to compute response maps of all facial landmark points. Such a data-driven method could make full use of holistic information in a facial image for global estimation of facial landmarks. After that, the maximum points in the response maps are fitted with a pre-trained Point Distribution Model (PDM) to generate the initial facial shape. This model-driven method is able to correct the inaccurate locations of outliers by considering the shape prior information. Finally, a weighted version of Regularized Landmark Mean-Shift (RLMS) is employed to fine-tune the facial shape iteratively. This Estimation-Correction-Tuning process perfectly combines the advantages of the global robustness of data-driven method (FCN), outlier correction capability of model-driven method (PDM) and non-parametric optimization of RLMS. Results of extensive experiments demonstrate that our approach achieves state-of-the-art performances on challenging datasets including 300W, AFLW, AFW and COFW. The proposed method is able to produce satisfying detection results on face images with exaggerated expressions, large head poses, and partial occlusions.
\end{abstract}


\IEEEpeerreviewmaketitle

%%%%%%%%%%%%%%%%%
%% Introduction
%%%%%%%%%%%%%%%%%
\section{Introduction}
\label{Introduction}

\IEEEPARstart{F}{acial} landmark detection, or face alignment, aims to localize a set of semantic points, such as eye-corners, nose tip, and lips, on a face image accurately and efficiently. It is a fundamental problem in computer vision study with wide applications in face recognition, facial expression analysis, human-computer interaction, video games,~\etc. Impressive progress has been made on facial landmark detection in recent years and current methods could provide reliable results for near-frontal face images~\cite{cao2014face,xiong2013supervised,ren2014face,zhang2014coarse,zhu2015face,zhang2016learning}. However, it is still a challenging problem for localizing landmarks in face images with partial occlusions or large appearance variations due to illumination conditions, poses, and expression changes.

In order to deal with these difficulties in facial landmark detection, a robust solution should make use of both textural appearance information in facial images and the inherent structural constraint of facial landmarks.

For appearance information,
%the classical method Active Shape Models (ASMs) construct the appearance model individually for each facial points,
the classic Active Shape Models (ASMs)~\cite{cootes1992active} build the profile models via multivariate Gaussians to capture the appearance variance.
Active Appearance Models (AAMs)~\cite{cootes2001active,tzimiropoulos2013optimization} construct the holistic appearance on the shape-free warped textures.
%and Active Appearance Models (AAM) builds the holistic appearance model by applying PCA onto the shape-free warped textures.
%Though AAM can achieve high accurate results reported in recent literatures, it's hard for them to cope with occlusions.
Both ASMs and AAMs build generative models based on the appearance information, and it is hard for them to reflect complex and subtle face variations in natural conditions.
%Hence they can not generalize to unseen faces, and suffer from the high dimension of appearance parameters when trained on facial images with large variations.
%To better handle the large scale of training data,
In order to capture the textural information more effectively,
Constrained Local Models (CLMs)~\cite{cristinacce2006feature,saragih2011deformable} train the local experts discriminatively for each facial landmark.
%The local experts are classically based on SVM classifier, and other forms of experts are also exploited in literatures.
%The response maps from local experts indicate the likelihood that the points in local patches align to the corresponded landmark.
%Moreover, regression-based methods directly use the shape-indexed feature to regress the position of facial landmarks.
For regression-based methods~\cite{cao2014face,xiong2013supervised,ren2014face}, the appearance information is encoded in shape-indexed features~\cite{cao2014face} represented by handcrafted descriptors (\eg SIFT)~\cite{xiong2013supervised} or learned in a data-driven manner~\cite{ren2014face}. Meanwhile, Deep Neural Networks (DNNs) also prove to be competent for extracting features invariant to various undesired conditions~\cite{sun2013deep,zhang2016learning,bulat2016convolutional,trigeorgis2016mnemonic}.
Different network architectures including Convolutional Neural Networks (CNNs)~\cite{sun2013deep}, Auto-Encoder~\cite{zhang2014coarse,li2017fast}, Recurrent Neural Networks (RNNs)~\cite{xiao2016robust,trigeorgis2016mnemonic,lai2016deep} and Fully Convolutional Networks (FCNs)~\cite{bulat2016convolutional,Bulat2017HowFar} have been exploited as regressors to predict the facial landmark shapes or response maps for each individual facial landmark.
Data-driven methods have shown their expressive power in handling the appearance variations caused by head poses and expressions~\cite{zhang2016learning,lv2017deep,Bulat2017HowFar}.

For shape constraint, ASMs, AAMs and CLMs build Point Distribution Model (PDM)~\cite{cootes1992active} by applying Principal Component Analysis (PCA) onto training shapes normalized via Procrustes analysis. In this way, the structural relationships between landmarks are embedded in the PCA bases of PDM. The shape model PDM could be viewed as a global constraint imposed on the local appearance models in ASMs or CLMs. In the last decade, more complex models, such as Markov random field~\cite{valstar2010facial} and tree-structured models~\cite{zhu2012face}, are proposed to capture the dependencies between facial landmarks. Unfortunately, the practical usages of these models are limited due to inefficient inference or the lack of loopy spatial constraints.
%For instance, Markov random field is exploited to restrict the search of the landmarks shapes under a graph model.
%To alleviate the optimization complexity, tree-structured models are used to define the relation between neighboring landmarks.
%Though the inference for the tree-structured model is more efficient, its design is task-specified and prone to unreasonable landmark shapes.
For regression-based methods like~\cite{cao2014face,ren2014face}, the shape constraint is encoded implicitly in the cascaded regressors, which achieves breakthroughs in accuracy and speed of facial landmark detection algorithms. However, it is still difficult for them to cover a wide range of head poses~\cite{xiong2013supervised}.
Moreover, deep learning based methods~\cite{bulat2016convolutional,xiao2016robust,lv2017deep} typically model the dependencies between landmarks by resorting to complex network structures, which may lead to deeper architectures or sophisticated learning strategies.

In this paper, a novel three-step framework named ECT (Estimation-Correction-Tuning) is proposed for facial landmark detection by combining data- and model-driven methods.
In the estimation step, the appearance information is captured discriminatively in a data-driven manner.
%dense predictions for each landmark are outputted by the regressor.
Correction is then performed to impose the structural constraint on the estimated shape by a generative shape model.
% In the correction step, a generative shape model capturing the structural constraints is imposed to regulate the shape estimated from the previous step.
The resultant shape is then fine-tuned iteratively to balance the data-driven and model-driven efforts.
% In the tuning step, the estimated shape is fine-tuned iteratively to balance the data-driven and model-driven efforts.
The merits of the proposed method are emphasized as follows.
%The novelty and contributions of this paper are summarized as follows.

(a) Data-driven estimation of initial landmarks: A Fully Convolutional Network (FCN) is employed to learn a desirable response map for each landmark. The ideal response map is defined as a 2D Gaussian with its center located precisely at the facial landmark point~\cite{pfister2015flowing,bulat2016convolutional}. Such a data-driven method could make full use of holistic information to avoid local minimum traps.

(b) Model-driven correction of outliers: The initial landmarks extracted from the response maps are fitted into a pre-trained Point Distribution Model (PDM), so that the outliers could be corrected according to the shape prior. Moreover, utilizing the structural information is beneficial to inferring invisible landmarks in cases with partial occlusions.

(c) Non-parametric fine-tuning of facial shape: The confidence of the response map is integrated into a weighted version of Regularized Landmark Mean-Shift (RLMS)~\cite{saragih2011deformable} framework. The combination of the evidence from the response maps and parametric shape prior further improves the accuracy of facial landmark detection. Both the expressive power of data-driven method (FCN) and the reasonable shape constraints (PDM) are incorporated into a general framework for robust facial landmark detection.

(d) The novel pipeline makes full use of each component and contributes to the success of our approach. Extensive experiments show that the proposed method achieves state-of-the-art performances on the in-the-wild face datasets including 300W~\cite{sagonas2013300}, AFLW~\cite{kostinger2011annotated}, AFW~\cite{zhu2012face} and COFW~\cite{burgos2013robust}. The success of the Estimation-Correction-Tuning strategy and the idea of combining data- and model-driven methods provide a novel solution to develop more advanced methods for robust facial landmark detection. Moreover, our method also has great potential in solving other computer vision problems, such as human pose estimation and image segmentation.

The rest of this paper is organized as follows.
Section~\ref{RelatedWork} briefly reviews representative works related to our method.
The technical details of our method are presented in Section~\ref{Approach}.
Experimental results are reported in Section~\ref{Experiment}.
Finally, Section~\ref{Conclusion} concludes this paper.

\begin{figure*}[t]
\begin{center}
   \includegraphics[height=70mm]{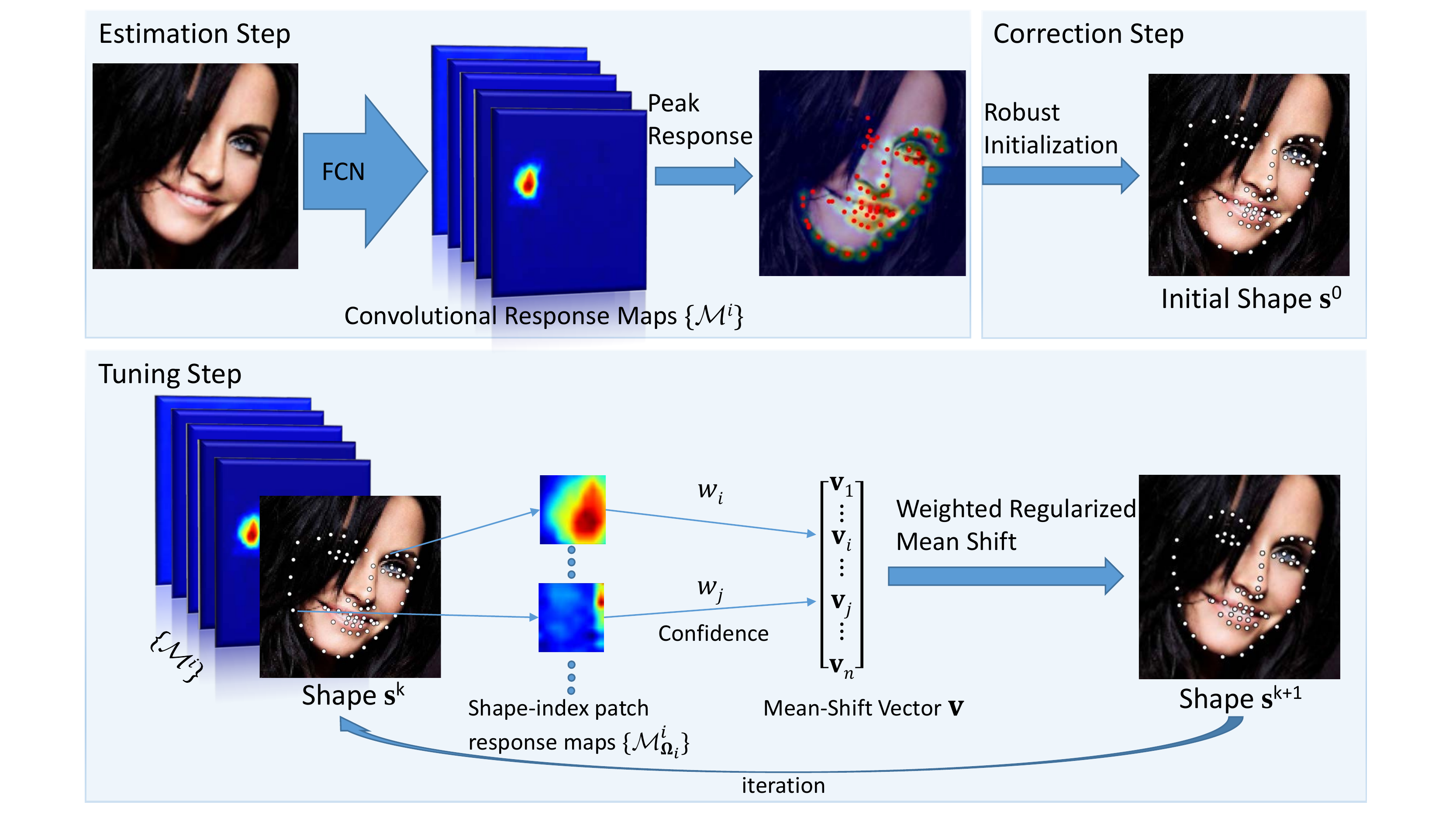}
   \caption{Overview of the proposed ECT (Estimation-Correction-Tuning) approach for facial landmark detection.
   The framework mainly consists of three steps:
   1) The Estimation Step obtains a coarse landmark shape according to the peak response points in the response maps regressed by a fully convolutional network;
   2) The Correction Step fits the coarse estimation with the PDM to get a more reasonable initial shape for subsequent procedures;
   3) The Tuning Step fine-tunes the landmark shape iteratively based on the proposed weighted regularized mean shift.
   }
\label{fig:flowchart}
\end{center}
\end{figure*}

%%%%%%%%%%%%%%%%%
%% Related Work
%%%%%%%%%%%%%%%%%
\section{Related work}
\label{RelatedWork}

%According to how the image features are utilized,
Facial landmark detection algorithms could be roughly classified into three categories, part-based methods~\cite{tzimiropoulos2014gauss,zhu2012face}, holistic methods~\cite{cootes2001active,tzimiropoulos2013optimization}, and regression-based methods~\cite{cao2014face,xiong2013supervised,ren2014face,zhang2014coarse,bulat2016convolutional,lv2017deep}.
The part-based methods assemble the outputs from part models and refine the results jointly under certain regulations.
The holistic methods manage to optimize a holistic appearance model that best fits the given test face.
% For regression-based methods, the shapes are typically regressed directly from the shape-index features.
For regression-based methods, the regressors are typically built to regress the landmark shape directly or predict response maps for each landmark.

\subsection{Part-based Methods}
Our method has an inherent relationship with a sort of part-based methods~\cite{cootes1992active,cristinacce2006feature,saragih2011deformable},
%Those methods typically assemble the outputs from part models and refine the results overall under certain regulations.
which typically build a local model for each landmark to suggest the update direction and use a shape model to regulate the result globally.
%The local models capture the appearance information around the landmarks, and the shape model encodes the structure relationship between them.
This idea originates from the famous ASM~\cite{cootes1992active} and is extended in CLM~\cite{cristinacce2006feature} to build the more discriminative patch experts.
%For each landmark, ASM builds the profile models via multivariate Gaussians to capture the appearance variance.
%which is classically limited in a neighbourhood line orthogonal to the shape boundary.
%CLM extends the profile models in ASM to the more discriminative local experts and approximates its response maps parametrically during the fitting period.
%For each facial landmark, it trains an SVM-like classifier and calculates a response map indicated the likelihood that position aligns to the landmark.
%In the fitting period, it approximates the response maps using parametrical models such as Gaussian mixtures so that the update has a closed-form.
ASM and CLM establish a general framework for facial landmark detection.
%and many improvements of patch experts and optimization method have been proposed in recent years~\cite{saragih2011deformable,yu2013pose,asthana2013robust}.
Different patch experts and more efficient optimization methods have been proposed in recent years~\cite{saragih2011deformable,yu2013pose,asthana2013robust}.
Saragih \etal~\cite{saragih2011deformable} interpret the CLM in a probabilistic perspective and propose the well-known RLMS via a nonparametric approximation of the response maps.
Asthana \etal~\cite{asthana2013robust} compress the response maps using dimensionality reduction and regress the shape parameters from the low-dimensional representation of response maps.
%They proposed a non-parametric approximation to the response maps via Gaussian kernel estimation (KDE), and updated the shape iteratively using the E-M algorithm.
%Baltrusaitis \etal used Local Neural Field (LNF) as patch experts, which is more discriminative than SVM-like classifiers.
Baltrusaitis \etal~\cite{baltrusaitis2013constrained} enhance the patch experts using Local Neural Field (LNF) and propose a non-uniform version of RLMS by considering the reliability of patch experts.
Recently, they~\cite{zadeh2017convolutional} also introduce Convolutional Experts Network (CEN) to replace LNF and achieve higher performances.
The weighted term in their non-uniform RLMS is pre-determined and fixed during testing, which is different from our formulation.
%Part-based methods are relatively easier to optimize and robust to partial occlusions.
One limitation of the above mentioned part-based methods is that the utilization of the appearance information is limited in the vicinity of each part.
To alleviate this, Alabort-i-Medina \etal~\cite{alabort2015unifying} propose a unified formulation to merge the holistic and part-based methods.
Alternately, we extend the patch experts to make full use of holistic information and utilize the confidence of each expert explicitly to handle the occlusions.

\subsection{Regression-based Methods}
Tackling facial landmark detection as a regression problem is another trend in recent years~\cite{cao2014face,ren2014face,sun2013deep,zhang2014coarse}.
Typical approaches~\cite{cao2014face,xiong2013supervised} use cascaded regressors to predict the coordinates of landmarks directly from shape-indexed features.
More recent methods~\cite{zhang2014coarse,lv2017deep,xu2017joint} develop strategies such as coarse-to-fine prediction and global-to-local regression to capture information at different scales.
On the other hand, Tompson \etal~\cite{tompson2014joint} argue that mapping directly from image features to coordinates of joints is highly non-linear and difficult in the context of human pose estimation.
Instead of regressing the coordinates, they propose to regress heat-maps, which indicates the per-pixel likelihood of alignment for each body joint.
%To exploit the structural constraints, they built a Spatial-Model to pass the message between joints.
Since then, predicting dense heat-maps via FCN-based networks is exploited for both human pose estimation~\cite{pfister2015flowing,newell2016stacked} and facial landmark detection~\cite{huang2015densebox,bulat2016convolutional,yang2017stacked,Bulat2017HowFar}.
In~\cite{pfister2015flowing}, Pfister \etal propose to fuse the shallower and deeper layers of the convolutional network to learn spatial dependencies between body parts.
In~\cite{newell2016stacked}, Newell \etal introduce Stacked Hourglass Network for human pose estimation by incorporating multi-resolution features to learn the spacial relationships between joints.
%Chu \etal~\cite{chu2016structured} introduce structured features so that the information between joints could flow at feature level according to a pre-defined bi-directional tree model.
For facial landmark detection,
Huang \etal~\cite{huang2015densebox} introduce a unified FCN framework named DenseBox for simultaneous landmark localization and face detection.
Bulat \etal~\cite{bulat2016convolutional} propose a two-step detection-followed-by-regression network to predict the detection scoremap for each landmark.
Xiao \etal~\cite{xiao2016robust} propose to extract shape-indexed deep features from FCN and refine the landmark locations recurrently via LSTM.
%a recurrent attentive-refinement network (RAR) which takes the shape-indexed deep features and the last information as inputs and recurrently refine the shapes.
In the recent works of~\cite{yang2017stacked,Bulat2017HowFar}, Stacked Hourglass Network is also shown to be effective for facial landmark detection.
Additionally, landmark heatmaps are used as indicators to help extract discriminative features from images.
In~\cite{Kowalski2017DAN}, Kowalski \etal revise Cascade Shape Regression based methods by stacking the landmark heatmaps with the raw image as input for each stage.
Kumar \etal~\cite{kumar2017kepler} also adopt a similar strategy in an iterative multi-task learning framework to predict the head pose, visibility and position of each landmark simultaneously.
The methods mentioned above either build the pre-defined structure model into the network or implicitly exploit the structural constraint by resorting to the elaborate network architectures.
%Xiao \etal developed a recurrent attentive-refinement network (RAR) which takes the shape-indexed deep features and the last information as inputs and recurrently refine the shapes.
In contrast, we propose a framework based on convolutional response maps and explicitly utilize the structure model to regulate the landmark locations with respect to the confidence of the response maps, which is shown to be effective in our experiments.

\section{Approach}
\label{Approach}

The framework of the proposed ECT approach is shown in Fig.~\ref{fig:flowchart}.
Given an input face image, the landmark detection result is obtained after all three steps of ECT, namely estimation, correction and tuning step.
%There are mainly three steps namely Estimation Step, Correction Step and Tuning Step to achieve a robust facial landmark detection result, given an input face image.
The Estimation Step aims to compute a global localization of the initial landmarks based on the peak response points in the response maps, which are learned from a fully convolutional network (FCN). After that, a more reasonable and accurate initial shape for subsequent procedures is computed by correcting the outlier landmarks using a pre-trained point distribution model (PDM). Finally, the landmark locations are fine-tuned based on the proposed weighted regularized mean shift.

This section firstly introduces the problem formulation for the proposed approach.
Then the principal parts of the ECT approach, namely convolutional response map, robust initialization and weighted regularized mean shift, are presented later in detail.

\begin{figure*}[t]
    \centering
    \begin{subfigure}[b]{0.67\textwidth}
        \centering
        \includegraphics[width=\textwidth]{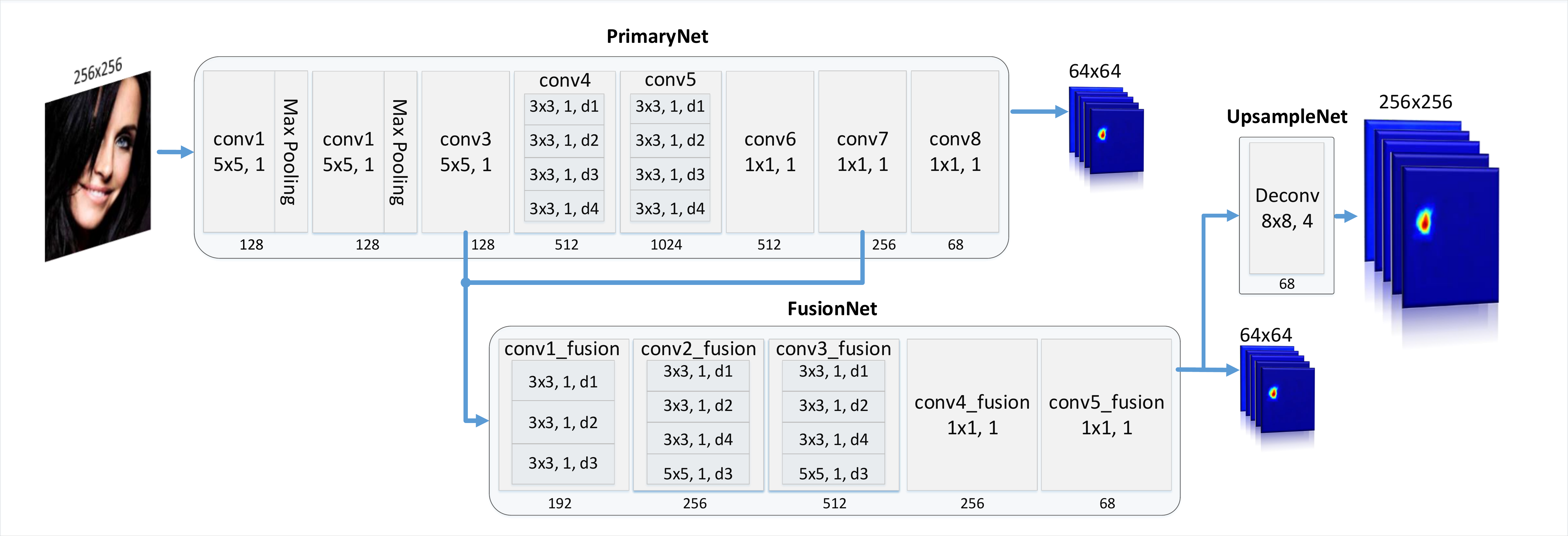}
        \caption{}
        \label{fig:FCNoverview}
    \end{subfigure}
    \begin{subfigure}[b]{0.32\textwidth}
        \centering
        \includegraphics[width=\textwidth]{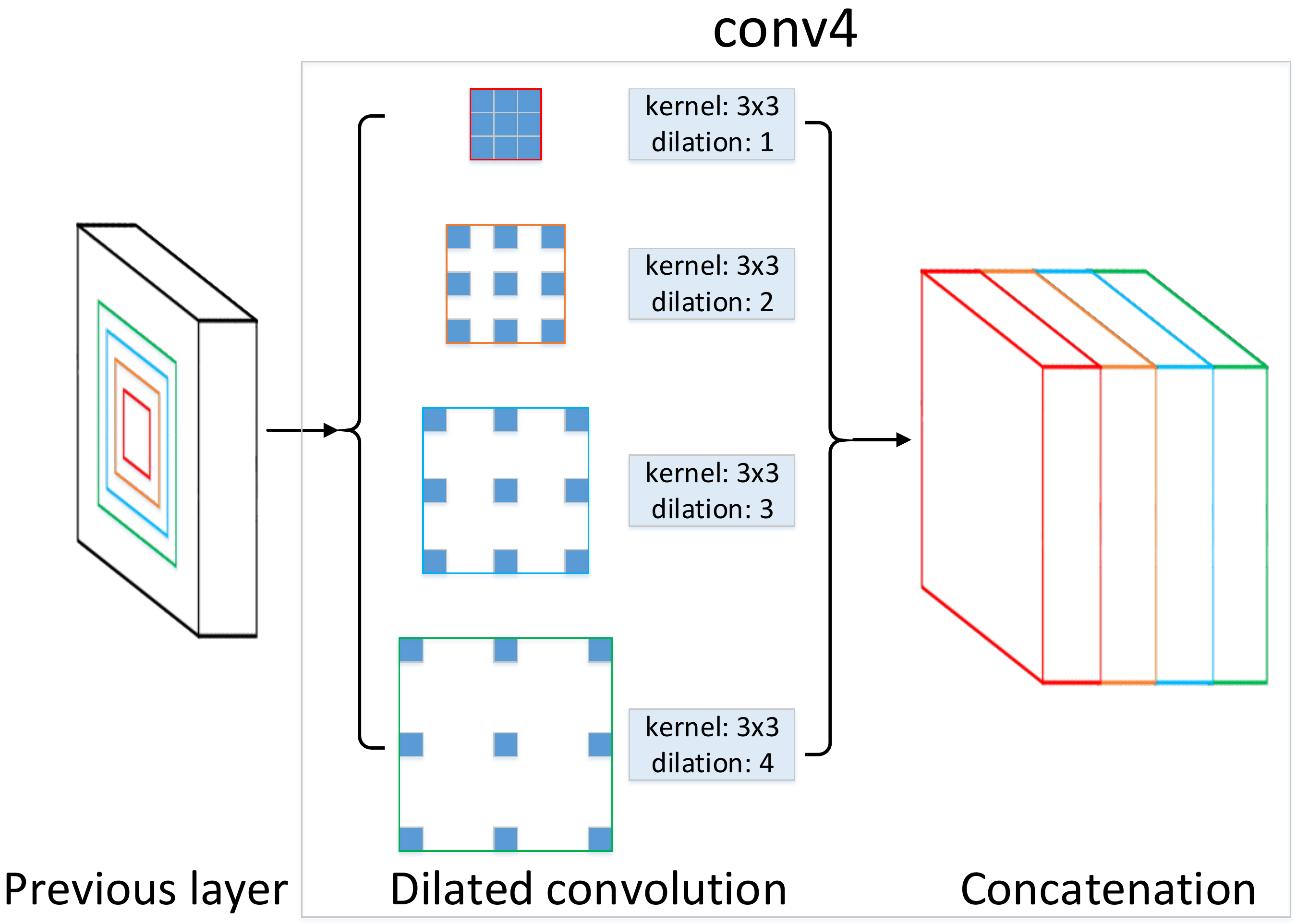}
        \caption{}
        \label{fig:Inception}
    \end{subfigure}
    \caption{Architecture of the proposed fully convolutional network (FCN).
    (a) Overview of our FCN. The filter parameter of each layer is denoted in the form of (kernel size, stride [, d*]) where d* means that there is a dilated convolution with the dilation factor of *. The number below the layer block shows the size of the output channel. ReLU nonlinearity is employed after each convolutional layer except conv8 and conv5\_fusion.
    (b) The Inception module with different dilated convolution filters. Note that dilated convolution with same number of filter parameters could have receptive fields with different sizes.}
    % Dilated convolution could have the same number of filter parameters but different size of receptive fields.}
\label{fig:FCNarch}
\end{figure*}

\subsection{Problem Formulation}
Point distribution model (PDM)~\cite{cootes1992active} is widely used in classic part-based methods.
It models the shape with global rigid transformations (scaling, in-plane rotation, and translation) as well as non-rigid variations (head poses and expressions).
%which can be regarded as modeling the similarity transform components and the head pose and expression respectively.
A 2D shape $\mathbf{s}=[{{\mathbf{x}}_{1}^\mathsf{T}},\ldots, {{\mathbf{x}}_{n}^\mathsf{T}}]^\mathsf{T}\in {{\mathbb{R}}^{2n\times 1}}$ could be considered as the concatenation of $n$ landmarks, with coordinates of the $i$-{th} landmark written as ${\mathbf{x}}_{i}=[{x}_{i}, {y}_{i}]^\mathsf{T}$. Further decomposition of ${\mathbf{x}}_{i}$ could be expressed in the following equation:
\begin{equation}
\label{eq:PDM}
\mathbf{x}_i=s\mathbf{R}(\overline{{\mathbf{x}}}_i+{\mathbf{\Phi}_i \mathbf{q}})+\mathbf{t}
\end{equation}
where $\mathbf{p}=\{s,\mathbf{R},\mathbf{t},\mathbf{q}\}$ denotes the parameters of PDM. Concretely, $\mathbf{p}$ consists of a set of global rigid transform parameters (global scaling $s$, rotation $\mathbf{R}$, translation $\mathbf{t}$) and non-rigid parameters $\mathbf{q}$.
$\overline{\mathbf{x}}_i$ and $\mathbf{\Phi}_i$ denote the pertaining sub-matrix of the $i$-{th} landmark in the mean shape $\overline{\mathbf{s}}$ and the shape components $\mathbf{\Phi}$, respectively.
$\mathbf{\Phi}$ is the collection of eigenvectors corresponding to the $m$ largest eigenvalues by applying PCA to a set of training shapes.
Given enough training samples, $\mathbf{\Phi}$ is capable of encoding rich expressions.
%Given enough training samples, $\mathbf{\Phi}$ can encode the rich expression adequately.
Assuming that the rigid transformation mentioned earlier has a non-informative prior and that the non-rigid shape parameter $\mathbf{q}$ exhibits Gaussian distributions, the PDM parameter $\mathbf{p}$ has the following prior:
\begin{equation}
\label{eq:PDMprior}
p(\mathbf{p})\varpropto\mathcal{N}(\mathbf{q};\mathbf{0},\mathbf{\Lambda});~~\mathbf{\Lambda }=\text{diag }\{[{{\lambda }_{1}},\ldots, {{\lambda }_{m}}]\}
\end{equation}
where ${\lambda}_{i}$ denotes the eigenvalue of the $i$-{th} eigenvector in ${\mathbf{\Phi}}$.
With this prior knowledge, the parameter $\mathbf{p}$ could be inferred in a Bayesian manner.
Assuming the detection results are conditionally independent for each landmark,
the posterior distribution of $\mathbf{p}$ could then be written as:
\begin{equation}
\label{eq:poster}
p(\mathbf{p}|\{{{l}_{i}}=1,{ \mathcal{D}_{i}}\}_{i=1}^{n},\mathcal{I})\propto p(\mathbf{p})\prod\limits_{i=1}^{n}{p({{l}_{i}}=1|{{\mathbf{x}}_{i}},{\mathcal{D}_{i}},\mathcal{I})}
\end{equation}
where ${{l}_{i}}\in \{1,-1\}$ indicates whether the $i$-{th} landmark is aligned or misaligned on coordinate ${\mathbf{x}}_{i}$ for image $\mathcal{I}$,
and ${\mathcal{D}}_i$ is the given detection expert for the $i$-{th} landmark.

Assuming that there is a set of candidate coordinates ${{\mathbf{\Psi }}_{i}}$ for the $i$-{th} landmark,
the conditional likelihood $p({{l}_{i}}=1|{{\mathbf{x}}_{i}},{\mathcal{D}}_i,\mathcal{I})$ could be approximated with a nonparametric representation~\cite{saragih2011deformable} using Kernel Density Estimator (KDE).
Mathematically, the conditional likelihood has the following form:
\begin{equation}
\label{eq:conditionalLikelihood}
p({{l}_{i}}=1|{{\mathbf{x}}_{i}},{\mathcal{D}}_i,\mathcal{I})=\sum\limits_{{{\mathbf{y}}_{i}}\in {{\mathbf{\Psi }}_{i}}}{{{\pi }_{{{\mathbf{y}}_{i}}}}\mathcal{N}({{\mathbf{x}}_{i}};{{\mathbf{y}}_{i}},{\rho}_i \mathbf{I})}
\end{equation}
where ${{\pi }_{{{\mathbf{y}}_{i}}}}=p({{l}_{i}}=1|{{\mathbf{y}}_{i}},{\mathcal{D}}_i,\mathcal{I})$ is corresponding to the response map (normalized) to be introduced in the next subsection.
${\rho}_i=\frac{{\rho}^2}{w_i}$ is used to smooth the response map, where $\rho$ is a free parameter and $w_i$ adjusts the smoothness according to the confidence of the detection expert ${\mathcal{D}}_i$ for image $\mathcal{I}$.

Combining Eqs.~\eqref{eq:poster} and~\eqref{eq:conditionalLikelihood} gives us the following expression:
\begin{equation}
\label{eq:posterFinal}
%\begin{split}
p(\mathbf{p}|\{{{l}_{i}}=1,{{\mathcal{D}}_{i}}\}_{i=1}^{n},\mathcal{I})\propto p(\mathbf{p})\prod\limits_{i=1}^{n}{\sum\limits_{{{\mathbf{y}}_{i}}\in {{\mathbf{\Psi }}_{i}}}{{{\pi }_{{{\mathbf{y}}_{i}}}}\mathcal{N}({{\mathbf{x}}_{i}};{{\mathbf{y}}_{i}},{\rho}_i \mathbf{I})}}
%\end{split}
\end{equation}
%where $w_i=p({{\mathcal{D}}_{i}}|\mathcal{I})$.

Eq.~\eqref{eq:posterFinal} could be solved iteratively using the EM algorithm and the mean-shift algorithm~\cite{saragih2011deformable}.
In the E-step, the posterior over $\mathbf{y}_i$ could be evaluated as follows when treating the candidates $\{\mathbf{y}_i\}_{i=1}^{n}$ as hidden variables:
\begin{equation}
\label{eq:Estep}
{{w}_{{{\mathbf{y}}_{i}}}}=p({{\mathbf{y}}_{i}}|{{l}_{i}}=1,{{\mathbf{x}}_{i}},{{\mathcal{D}}_{i}},\mathcal{I})=\frac{{{\pi }_{{{\mathbf{y}}_{i}}}}\mathcal{N}\left( {{\mathbf{x}}_{i}};{{\mathbf{y}}_{i}},{\rho}_i \mathbf{I} \right)}{\sum\nolimits_{{{\mathbf{z}}_{i}}\in {{\mathbf{\Psi }}_{i}}}{{{\pi }_{{{\mathbf{z}}_{i}}}}\mathcal{N}\left( {{\mathbf{x}}_{i}};{{\mathbf{z}}_{i}},{\rho}_i \mathbf{I} \right)}}
\end{equation}
Then, the M-step works on minimizing the Q function:
\begin{equation}
\label{eq:Mstep}
\begin{split}
Q(\mathbf{p})&={{E}_{q(\mathbf{y})}}[-\ln \{p(\mathbf{p})\prod\limits_{i=1}^{n}{p({{l}_{i}}=1,{{\mathbf{y}}_{i}}|{{\mathbf{x}}_{i}},{\mathcal{D}}_{i},\mathcal{I})}\}]\\
&\propto \left\| \mathbf{q} \right\|_{{{\mathbf{\Lambda }}^{-1}}}^{2}+\sum\limits_{i=1}^{n}{{{w}_{i}}\sum\limits_{{{\mathbf{y}}_{i}}\in {{\mathbf{\Psi }}_{i}}}{\frac{{{w}_{{{\mathbf{y}}_{i}}}}}{{\rho}^2 }{{\left\| {{\mathbf{x}}_{i}}-{{\mathbf{y}}_{i}} \right\|}^{2}}}}
\end{split}
\end{equation}
where $q(\mathbf{y})=\prod\nolimits_{i=1}^{n}{p({{\mathbf{y}}_{i}}|{{l}_{i}}=1,{{\mathbf{x}}_{i}},{{\mathcal{D}}_{i}},\mathcal{I})}$.
The iterative solution $\Delta \mathbf{p}$ for each update could thus be written as:
\begin{equation}
\label{eq:pUpdate}
\Delta \mathbf{p}=-{{({\rho}^2 {{\widetilde{\mathbf{\Lambda }}}^{-1}}+{{\mathbf{J}}^\mathsf{T}}\mathbf{WJ})}^{-1}}({\rho}^2 {{\widetilde{\mathbf{\Lambda }}}^{-1}}\mathbf{p}-{{\mathbf{J}}^\mathsf{T}}\mathbf{Wv})
\end{equation}
In the above equation, $\widetilde{\mathbf{\Lambda }}=\text{diag}\{[\mathbf{0},{{\lambda }_{1}},\ldots,{{\lambda }_{m}}]\}$, $\mathbf{J}=[\mathbf{J}_1,\ldots,\mathbf{J}_n]$, where $\mathbf{J}_i$ is the Jacobian of PDM in Eq.~\eqref{eq:PDM}, $\mathbf{W}=\text{diag}\{[{{w}_{{{x}_{1}}}},{{w}_{{{y}_{1}}}},\ldots,{{w}_{{{x}_{n}}}},{{w}_{{{y}_{n}}}}]\}$ with ${w}_{{{x}_{i}}}={w}_{{{y}_{i}}}={w}_i$, $\mathbf{v}=[{\mathbf{v}}_{1}^\mathsf{T},\ldots,{\mathbf{v}}_{n}^\mathsf{T}]^\mathsf{T}$ is the mean shift vectors of all landmarks:
\begin{equation}
\label{eq:meanshift}
{{\mathbf{v}}_{i}}=\left(\sum\limits_{{{\mathbf{y}}_{i}}\in {{\mathbf{\Psi }}_{i}}}{\frac{{{\pi }_{{{\mathbf{y}}_{i}}}}\mathcal{N}\left( \mathbf{x}_{{{\mathbf{y}}_{i}}}^{c};{{\mathbf{y}}_{i}},{\rho}_i \mathbf{I} \right)}{\sum\nolimits_{{{\mathbf{z}}_{i}}\in {{\mathbf{\Psi }}_{i}}}{{{\pi }_{{{\mathbf{z}}_{i}}}}\mathcal{N}\left( \mathbf{x}_{i}^{c};{{\mathbf{z}}_{i}},{\rho}_i \mathbf{I} \right)}}}{{\mathbf{y}}_{i}}\right)-\mathbf{x}_{i}^{c}
\end{equation}
where $\mathbf{x}_i^c$ is the currently estimated position of the $i$-{th} landmark.

Eqs.~\eqref{eq:pUpdate} and \eqref{eq:meanshift} demonstrate that our algorithm alternates between computing the move step from response maps and regularizing it with the shape model's constraint.
The main difference between our formulation and RLMS~\cite{saragih2011deformable} is that a weight value is assigned to each landmark mean-shift vector before it is projected onto the subspace spanned by the PDM's Jacobian, which contributes a key factor to the success in robust facial landmark detection.
Note that the weights are updated in each tuning step, which is different from the non-uniform RLMS~\cite{baltrusaitis2013constrained}.
Specifically, the mean-shift vector calculated from response maps is selectively projected onto the PCA space according to the latest weight matrix $\mathbf{W}$,
so that the tuning step could effectively reach the balance between the efforts from detection experts and the global prior information from PDM.

\subsection{Convolutional Response Map}
\label{methodFCN}

Regression-based methods~\cite{cao2014face,sun2013deep,xiong2013supervised} train regressors to predict the landmark location $\mathbf{x}_i$ directly.
Following the previous work~\cite{pfister2015flowing,bulat2016convolutional}, the regressor FCN in our method, namely the part detection expert, is used to regress the ideal response map for each landmark in a data-driven manner.
The ideal response map of the $i$-{th} landmark for image $\mathcal{I}$ is a single-channel image $\mathcal{M}^i$ with the same resolution as $\mathcal{I}$, and its pixel value at position $\mathbf{z}$ is defined as $\mathcal{M}_{\mathbf{z}}^{i}=\mathcal{N}(\mathbf{z};{{\mathbf{x}}_{i}^{*}},{{\sigma }^{2}}\mathbf{I})$, where $\mathbf{x}_{i}^*$ is the ground truth location of the $i$-{th} landmark, and $\sigma$ serves to control the scope of the response.

Fig.~\ref{fig:FCNoverview} shows an overview of the proposed FCN architecture. Our FCN network consists of three connected subnetworks, namely the PrimaryNet, FusionNet and UpsampleNet. Given an input image with the size of $256\times256$, the first two subnetworks regress the smaller response maps with the size of $64\times64$.
The last UpsampleNet is simply a deconvolutional layer which bilinearly upsamples the feature maps back to the size of the input image.

Given the training dataset $N=\{(\mathcal{I},\mathbf{s}^*)\}$, where $\mathbf{s}^*$ is the ground truth shape embedded in image $\mathcal{I}$,
the objective of the regressor becomes estimating the network weights $\lambda$ that minimize the following L2 loss function:
\begin{equation}
\label{eq:objective}
\mathcal{L}(\lambda )=\sum\limits_{(\mathcal{I},{{\mathbf{s}}^{*}})\in N}{\sum\limits_{i}{{{\left\| {{\mathcal{M}}^{i}}-{{\phi }^{i}}(\mathcal{I},\lambda ) \right\|}^{2}}}}
\end{equation}
where ${\phi}^i(\mathcal{I},\lambda )$ is the output of the $i$-{th} channel of the regression network fed with the image $\mathcal{I}$.
The loss functions for the PrimaryNet and FusionNet have the same loss function as expressed in~\eqref{eq:objective} with downsampled spatial resolutions.

The design of the PrimaryNet and FusionNet is originated from the pose estimation networks~\cite{pfister2015flowing} and adapted to the case of facial landmark detection.
As mentioned in~\cite{pfister2015flowing}, the PrimaryNet can not learn the spatial dependencies of landmarks very well.
To address this problem, conv3 and conv7 are firstly concatenated and then fed to the FusionNet.
%Such a strategy generates satisfactory response maps from FusionNet.
%And the output of PrimaryNet is only slightly inferior to FusionNet since the loss of FusionNet is also backpropagated through PrimaryNet.
The main difference between the architectures of our subnetworks and the pose estimation networks~\cite{pfister2015flowing} is that we adopt the Inception module~\cite{szegedy2015going} with different dilated convolution filters~\cite{YuKoltun2016} for layer conv4, conv5, conv1\_fusion, conv2\_fusion and conv3\_fusion.
All these dilated convolution sub-layers are concatenated together so that the next layer could extract features from different scales simultaneously (see illustration in Fig.~\ref{fig:Inception}).
Such an improvement could achieve a comparable regression result with the size of the model reduced by half.
% Such an improvement could achieve a comparable regression result with 20 megabytes of model size reduced, which accounts for more than 50 percentage of the whole model size in our case.
The benefit of using dilated convolution is significant, as it supports exponential expansion of the receptive field with the number of parameters growing linearly. These dilated convolution filters are well-suited for landmark detection task which requires pixel level texture information in multiple scales.

\begin{figure}[t]
    \begin{center}
        \foreach \idx in {2,3,4,5} {
            \begin{subfigure}[h]{0.11\textwidth}
                \centering
                \includegraphics[width=1.1\textwidth]{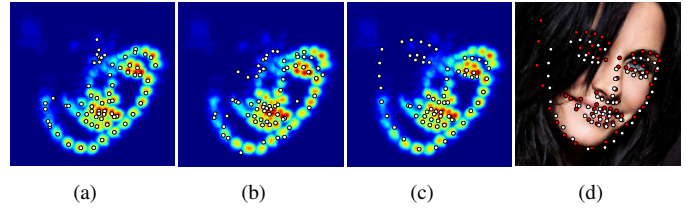}
                \caption{ }
            \end{subfigure}
        }
    \end{center}
   \caption{Robust initialization. (a)-(c) show the result of locating the landmarks according to the peak responce points, projecting the landmarks onto the PDM space, and correcting them using non-uniform regularization to obtain the initial shape, respectively.
   The response maps of all landmarks are superimposed over each other for better visualization.
   (d) shows the provided ground truth (red) and the initial shape (white) together with the face image.}
\label{fig:RobustInit}
\end{figure}

\begin{figure*}[t]
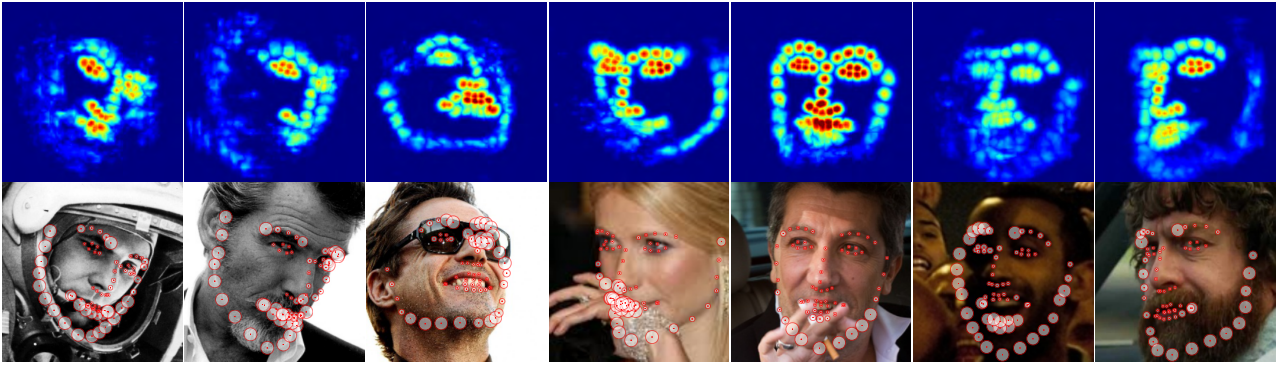

    \begin{center}
        \foreach \idx in {2,3,1,4,5,6,7} {
            \begin{subfigure}[h]{0.12\textwidth}
                \centering
                \includegraphics[width=1.1\textwidth]{Img/landmarkweight/rspmap\idx.pdf}
            \end{subfigure}
        }
        \foreach \idx in {2,3,1,4,5,6,7} {
            \begin{subfigure}[h]{0.12\textwidth}
                \centering
                \includegraphics[width=1.1\textwidth]{Img/landmarkweight/weightImg\idx.pdf}
            \end{subfigure}
        }
    \end{center}
   \caption{Visualization of the detection confidence of different facial landmarks. The upper row shows the superimposed response maps of all landmarks. The lower row shows the landmark position and the reliability calculated at the final stage. The radius of a landmark is inversely proportional to its reliability.}
\label{fig:landmarkweight}
\end{figure*}

\subsection{Robust Initialization}

At inferrence time, the image $\mathcal{I}$ is fed into a pre-trained FCN to obtain the response maps $\mathcal{M}=\{{\mathcal{M}}^i\}$.
The facial landmarks are firstly located at the peak response positions in the response maps and then fitted into the PDM to obtain a more reasonable and accurate landmark shape as initialization.
Generally, the coarsely estimated shape $\mathbf{s}$ could be regularized with the PDM directly by minimizing the reconstruction error.
However, such a regularization treats each landmark equally without considering their reliabilities.
For a more robust initialization, a non-uniform regularization is proposed by minimizing the following reconstruction error:
\begin{equation}
\underset{s,\mathbf{R},\mathbf{t},\mathbf{q}}{\mathop{\arg \min }}\,\left\| \mathbf{s}-s\widetilde{\mathbf{R}}(\overline{\mathbf{s}}+\mathbf{\Phi q})-\widetilde{\mathbf{t}} \right\|_{\mathbf{W}}^{2}
\end{equation}
where $\widetilde{\mathbf{R}}$ and $\widetilde{\mathbf{t}}$ are repeated copies of the rotation matrix $\mathbf{R}$ and translation vector $\mathbf{t}$ respectively, and $\mathbf{W}$ is a diagonal weight matrix where the diagonal elements are the weights of corresponding facial landmarks.
Note that the weight matrix is estimated in the same way as the $\mathbf{W}$ in Eq.~\ref{eq:pUpdate} and will be described in detail in the next subsection.
Adding the regularization term gives us the following objective function:
\begin{equation}
\underset{s,\mathbf{R},\mathbf{t},\mathbf{q}}{\mathop{\arg \min }}\,\left\| \mathbf{s}-s\widetilde{\mathbf{R}}(\overline{\mathbf{s}}+\mathbf{\Phi q})-\widetilde{\mathbf{t}} \right\|_{\mathbf{W}}^{2}+\gamma \left\| \mathbf{q} \right\|_{{{\mathbf{\Lambda }}^{-1}}}^{2}
\label{eq:regularized}
\end{equation}
where $\gamma$ is used to balance the two norms.

The optimal parameters of Eq.~\eqref{eq:regularized} could be obtained by iteratively solving the global similarity transform parameters $\{s,\mathbf{R},\mathbf{t}\}$ and non-rigid parameters $\mathbf{q}$.
Using the orthonormalization of the global similarity transform~\cite{matthews2004active}, the shape model (Eq.~\eqref{eq:PDM}) could be compactly written as:
\begin{equation}
\mathbf{s}=\overline{\mathbf{s}}+\mathbf{S}\mathbf{p}
\end{equation}
where $\mathbf{S}=[\mathbf{s}_{1}^{*},\ldots,\mathbf{s}_{4}^{*},\mathbf{\Phi }]\in {{\mathbb{R}}^{2n\times (4+m)}}$ is the concatenation of the similarity bases $\mathbf{s}_{i}^{*}$ with the original shape components $\mathbf{\Phi}$,
and $\mathbf{p}=[p_{1}^{*},\ldots,p_{4}^{*},\mathbf{q}^\mathsf{T}]^\mathsf{T}\in {{\mathbb{R}}^{(4+m)\times 1}}$ is re-defined as the concatenation of the similarity parameters $p_{i}^{*}$ with the non-rigid shape parameter $\mathbf{q}$.
Following the above decomposition, the optimal parameters of the initial shape ${\mathbf{s}}^{0}$ have the closed form:
\begin{equation}
{{\mathbf{p}}^{0}}={{(\gamma {{\widetilde{\mathbf{\Lambda }}}^{-1}}+{{\mathbf{S}}^\mathsf{T}}\mathbf{WS})}^{-1}}{{\mathbf{S}}^\mathsf{T}}\mathbf{Ws}
\label{eq:RobustInit}
\end{equation}
Note that Eq.~\eqref{eq:RobustInit} is equivalent to Eq.~\eqref{eq:pUpdate} when substituting $\mathbf{x}_{i}^{c}$ and $\mathbf{p}$ with $\mathbf{0}$ and regarding the move step $\mathbf{v}$ as $\mathbf{s}$.

Fig.~\ref{fig:RobustInit} demonstrates the effect of the robust initialization.
It could be seen that the proposed non-uniform regularization is able to correct the location error of outliers so that they could more accurately locate within the vicinity of the ground truth.
Such a correction step could provide a more reasonable initial shape for subsequent procedures and reduce the chance of failing during testing.
% reduce the failure cases occurred in testing.

\subsection{Weighted Regularized Mean Shift}

After computing the initial shape $\mathbf{s}^0$ from the correction step, the estimated shape is then fine-tuned iteratively using the weighted regularized mean shift.

In the $k$-{th} stage, the currently estimated shape $\mathbf{s}^k$ is obtained and the shape-index patch response maps are extracted from $\mathcal{M}$ for each landmark.
The shape-index patch response map $\{\pi_{\mathbf{y}_i}\}$ for the $i$-{th} landmark is a $r\times r$ square subarea %${\mathcal{A}_i}$
centered at $\mathbf{x}_i^k$ in the response map ${\mathcal{M}}^i$.
For those areas which are partially out of the range of ${\mathcal{M}}^i$ (\eg $\mathbf{x}_i^k$ is close to the boundary of ${\mathcal{M}}^i$ when $r$ is large enough), the missing parts are padded with zeros.
The collection of all coordinates in the square subarea for the $i$-{th} landmark is denoted as $\mathbf{\Omega}_i$.
Inspired by the work in~\cite{ren2014face}, the patch size $r$ is progressively shrinked from early stage to later stage.

For each shape-index patch response map, its confidence $w_i$ is estimated empirically as follows:
\begin{equation}
\label{eq:sigmoidweight}
{{w}_{i}}=sigmoid(a\frac{\sum\nolimits_{{{\mathbf{y}}_{i}}\in {{\mathbf{\Omega }}_{i}}}{{{\pi}_{{{\mathbf{y}}_{i}}}}}}{{{\operatorname{var}}_{{{\mathbf{y}}_{i}}\in {{\mathbf{\Omega }}_{i}}}}({\mathbf{y}_i},{{\pi}_{{{\mathbf{y}}_{i}}}})}+b)
\end{equation}
where ${\operatorname{var}}_{{{\mathbf{y}}_{i}}\in {{\mathbf{\Omega }}_{i}}}({\mathbf{y}_i},{{\pi}_{{{\mathbf{y}}_{i}}}})$ denotes the variance of the patch response map, $a$ and $b$ are two empirical parameters which could be optimized via cross-validation.
Eq.~\eqref{eq:sigmoidweight} suggests that the confidence of a local expert is proportional to the sum of its response values and inversely proportional to the degree of dispersion of its spatial distribution. Sigmoid function is used to normalize the confidence values within the range of $(0,1)$.
Fig.~\ref{fig:landmarkweight} visualizes the detection confidence of different facial landmarks on several examples.
It could be observed that those blurred or invisible parts are able to be differentiated from others easily.
%Noted that the weight in the initial step is similar to Eq.~\eqref{eq:sigmoidweight} where the patch response map is substituted with the full response map.

After calculating the weights, the mean shift vectors $\mathbf{v}_i$ are computed for all landmarks using Eq.~\eqref{eq:meanshift}, where the candidate coordinates ${{\mathbf{\Psi }}_{i}}$ are set equal to ${\mathbf{\Omega}_i}$,
and the response map $\{\pi_{\mathbf{y}_i}\}$ is normalized so that $\sum\nolimits_{{{\mathbf{y}}_{i}}\in {{\mathbf{\Psi }}_{i}}}{{{\pi}_{{{\mathbf{y}}_{i}}}}}=1$.
The confidence $w_i$ is assigned to each mean shift vector $\mathbf{v}_i$, and the update $\Delta \mathbf{p}$ for PDM parameters could be computed with the projection in Eq.~\eqref{eq:pUpdate}. Finally the latest estimated shape $\mathbf{s}^{k+1}$ is obtained for the next iteration by applying the incremental version of Eq.~\eqref{eq:PDM}.

The fine-tuning process could converge to stable outcomes after a number of iterations. A desirable result could typically be achieved within 5 iterations in our experiments. The complete process of our algorithm is summarized in Algorithm~\ref{alg:methodsummary}.

\begin{algorithm}[t]
\algrenewcommand\algorithmicrequire{\textbf{Require:}}
\algrenewcommand\algorithmicensure{\textbf{Output:}}
\algnewcommand{\LineComment}[1]{\State \(\triangleright\) #1}
\caption{ ECT (Estimation-Correction-Tuning) for facial landmark detection.}
\label{alg:methodsummary}
\begin{algorithmic}
\Require \\
The pre-trained FCN and PDM,
the input face image $\mathcal{I}$
\Ensure  ~\\
The facial shape $\mathbf{s}$
\newline \Comment {\textbf{Estimation step }}
\end{algorithmic}

\begin{algorithmic}[1]
\State Feed $\mathcal{I}$ into FCN and get the response maps $\{\mathcal{M}^i\}$
\State Obtain the coarse shape by locating the landmarks at the peak response positions in $\{\mathcal{M}^i\}$

\end{algorithmic}
\begin{algorithmic}
\LineComment {\textbf{Correction step}}
%\newline
\end{algorithmic}
\begin{algorithmic}[1]
\setcounter{ALG@line}{2}
\State Regularize the coarse shape with PDM and get the initial shape $\mathbf{s}^0$ using Eq.~\eqref{eq:RobustInit}
\end{algorithmic}
\begin{algorithmic}
\LineComment {\textbf{Tuning step}}
%\newline
\end{algorithmic}
\begin{algorithmic}[1]
\setcounter{ALG@line}{3}
\For{$k = 0~to~K$}
\For{the $i$-{th} landmark}
\State Calculate shape-index patch coordinates $\mathbf{\Omega}_i$
%\State Extract the patch response maps $\{{\pi}_{\mathbf{y}_i}=\mathcal{M}_{\mathbf{\Omega}_i}^i\}$
\State Extract patch response map $\{{\pi}_{\mathbf{y}_i}=\mathcal{M}_{\mathbf{y}_i}^i\}$ where ${{{\mathbf{y}}_{i}}\in {{\mathbf{\Omega }}_{i}}}$
%\State Calculate $\mathbf{v}_i$ and $w_i$ from $\{\mathcal{M}_{\mathbf{\Omega}_i}^i\}$, using Eqs.~\eqref{eq:meanshift} and~\eqref{eq:sigmoidweight} respectively
\State Calculate $w_i$ and $\mathbf{v}_i$ from $\{{\pi}_{\mathbf{y}_i}\}$, using Eqs.~\eqref{eq:sigmoidweight} and~\eqref{eq:meanshift} respectively
\EndFor
\State Calculate $\Delta \mathbf{p}$ using Eq.~\eqref{eq:pUpdate}
\State Update PDM parameters $\mathbf{p}^{k+1} = \mathbf{p}^{k} + \Delta \mathbf{p}$
\State Update shape $\mathbf{s}^{k+1}$ using Eq.~\eqref{eq:PDM}
\EndFor
\end{algorithmic}
\end{algorithm}

\section{Experiments}
\label{Experiment}

\textbf{Datasets.}
%The evaluation experiments involve two in-the-wild datasets, namely 300W and AFLW.
The evaluation experiments are conducted on four benchmark datasets, 300W~\cite{sagonas2013300}, AFLW~\cite{kostinger2011annotated}, AFW~\cite{zhu2012face} and COFW~\cite{burgos2013robust}, to demonstrate the effectiveness of the proposed method on challenging face images in natrual scenes.
\begin{itemize}
\item \textbf{300W}~\cite{sagonas2013300}: The 300W dataset contains near-frontal face images in the wild and provides 68 annotated points for each face.
To keep consistent with previous work, the datasets are partitioned and renamed as follows.
\emph{The 300W training set} contains 3,148 training images from AFW~\cite{zhu2012face}, LFPW~\cite{belhumeur2013localizing} and HELEN~\cite{le2012interactive}.
\emph{The common subset} of 300W contains 554 test images from LFPW and HELEN.
\emph{The challenging subset} of 300W contains 135 test images from IBUG~\cite{sagonas2013300}.
\emph{The fullset} of 300W is the union of the common and challenging subset.
\emph{The 300W test set} contains 600 test images which are provided officially by the 300W competition~\cite{sagonas2013300} and said to have a similar distribution to the IBUG dataset.

\item \textbf{AFLW}~\cite{kostinger2011annotated}: The original AFLW dataset contains about $25k$ in-the-wild faces with a wide range of head pose and provides up to 21 annotated points visible on each face. A subset of AFLW with a balanced distribution of head pose is selected in~\cite{jourabloo2015pose} and is denoted as AFLW-PIFA. Additional 13 landmarks are labeled later in~\cite{jourabloo2016large} so that 34 landmarks and their visible/invisible states are provided in AFLW-PIFA. For the evaluation on AFLW, 3,901 training images of AFLW-PIFA are used for training and the remaining 1,299 images are used for testing.
\item \textbf{AFW}~\cite{zhu2012face}: The AFW dataset is a popular benchmark for facial landmark detection, containing 468 faces of all pose ranges in 205 images.
A detection bounding box as well as up to 6 visible landmarks are provided for each face. The AFW dataset is only used for testing in our experiments due to the small number of samples.

\item \textbf{COFW}~\cite{burgos2013robust}: The COFW dataset contains in-the-wild face images with heavy occlusions, including 1345 face images for training and 507 face images for testing. For each face, 29 landmarks and the corresponding occlusion states are annotated in the COFW dataset.
\end{itemize}

\textbf{Evaluation metric.}
For fair comparison, the evaluation metrics are chosen as the common protocols in the literature~\cite{cao2014face,yu2013pose,sagonas2013300,zhu2016unconstrained,jourabloo2016large}.
The primary metric is the Normalized Mean Error (NME), which could be calculated as $\frac{1}{n}\sum\nolimits_{i=1}^{n}{\frac{{{\left\| {{\mathbf{x}}_{i}}-\mathbf{x}_{i}^{*} \right\|}_{2}}}{d}}$,
where $d$ denotes the normalized distance and $n$ is the number of facial landmarks involved in the evaluation.
Other evaluation metrics such as the Mean Average Pixel Error (MAPE), the Cumulative Error Distribution (CED) curve, the Area-Under-the-Curve (AUC) calculated from the CED curve and the failure rate are also reported in experiments for thorough analysis.

\textbf{Implementation details.}
The FCN in our experiments is implemented using the Caffe framework~\cite{jia2014caffe}. The FCN takes the input of a $256\times256$ face image and outputs a set of response maps with the same resolution.
To avoid overfitting, we randomly flip the input image horizontally and crop a $248\times248$ arbitrary sub-image from it. Then, we rotate it with a random angle from $-30^{\circ}$ to $30^{\circ}$ before rescaling it back to $256\times256$.
The variance $\sigma$  of the 2D Guassians in ideal response maps is set to $6$.
For those facial landmarks marked as invisible in the AFLW and COFW datasets, the ideal responses are re-defined as zeros instead of the 2D Gaussian responses.
During training of the FCN, the learning rate is fixed to $10^{-8}$ and the momentum is set to 0.95.
The parameter $a$ and $b$ in Eq.~\eqref{eq:sigmoidweight} is set to 0.25 and 25 respectively in our experiments.
The weighted regularized mean shift algorithm is implemented based on the Menpo project~\cite{alabort2014menpo}.
Code has been made publicly available.\footnote{https://github.com/HongwenZhang/ECT-FaceAlignment}

\begin{table}[t]
	\color{black}
  \centering
   \caption{Comparison of AUC (\%) and failure rate (\%) on the test set of the 300W competition. The results of other methods are obtained from~\cite{trigeorgis2016mnemonic,Kowalski2017DAN}.}
    \begin{tabular}{l|cc|cc}
    \toprule
          & \multicolumn{2}{c|}{51 points} & \multicolumn{2}{c}{68 points} \\
    %\midrule
    Method  & AUC   &  Failure rate & AUC   &  Failure rate \\
    \midrule
    ERT~\cite{kazemi2014one}   & 40.60  & 13.50  & 32.35  & 17.00  \\
    PO-CR~\cite{tzimiropoulos2015project} & 47.65  & 11.70  & -     & - \\
    %Chehra & 31.12  & 39.30  & -     & - \\
    SDM~\cite{xiong2013supervised} & 38.47  & 19.70  & -     & - \\
    CLNF~\cite{baltrusaitis2013constrained} & 37.65  & 17.17  & 19.55  & 38.83  \\
    %Face++~\cite{zhou2013extensive} & 53.29  & 5.33  & 32.81  & 13.00  \\
    %Yan \etal & 49.07  & 8.33  & 34.97  & 12.67  \\
    CFSS~\cite{zhu2015face}  & 50.79  & 7.80  & 39.81  & 12.30  \\
    MDM~\cite{trigeorgis2016mnemonic}   & 56.34  & 4.20  & 45.32  & 6.80  \\
    DAN~\cite{Kowalski2017DAN} & -  & -  & \textbf{47.00}  & \textbf{2.67}  \\
    \midrule
    ECT  & \textbf{58.26}  & \textbf{1.17}  & 45.98 & 3.17  \\
    \bottomrule
    \end{tabular}%
  \label{tab:compr300WTest}%
\end{table}%

\color{black}

\begin{table}[t]
  \color{black}
  \centering
  \caption{Comparison of NME (\%) with state-of-the-art methods on the fullset of 300W.}
  \begin{center}
    \begin{tabular}{lccc}
    \toprule
    Method & \begin{tabular}[x]{@{}c@{}} Common\\ Subset\end{tabular}
    &  \begin{tabular}[x]{@{}c@{}} Challenging\\ Subset\end{tabular} & Fullset \\
    \midrule
    CDM~\cite{yu2013pose}   &  10.10  &   19.54   &   11.94  \\
    TSPM~\cite{zhu2012face}  &  8.22  & 18.33 & 10.20 \\
    Smith \etal~\cite{smith2014nonparametric}  & -   &  13.30 & -\\
    DRMF~\cite{asthana2013robust} &6.65    &  19.79 & 9.22\\
    GN-DPM~\cite{tzimiropoulos2014gauss} &5.78  & -  & -  \\
    CLNF~\cite{baltrusaitis2013constrained} & -  & -  & 10.95 \\
    RCPR~\cite{burgos2013robust}  & 6.18  & 17.26 & 8.35 \\
    CFAN~\cite{zhang2014coarse}  & 5.50   & 16.78 & 7.69 \\
    ESR~\cite{cao2014face}   & 5.28  & 17.00    & 7.58 \\
    SDM~\cite{xiong2013supervised}   & 5.57  & 15.4  & 7.50 \\
    ERT~\cite{kazemi2014one}   & -     & -     & 6.40 \\
    LBF~\cite{ren2014face}   & 4.95  & 11.98 & 6.32 \\
    %Wu \etal~\cite{wu2015robust} & - & 11.52  & -  \\
    CFSS~\cite{zhu2015face}  & 4.73  & 9.98  & 5.76 \\
    TCDCN~\cite{zhang2016learning} & 4.80  & 8.60   & 5.54 \\
    3DDFA~\cite{zhu2016face}  & 5.53 & 9.56  & 6.31\\
    DDN~\cite{yu2016deep}   & -  &  -  &  5.65  \\
    RAR~\cite{xiao2016robust}   & \textbf{4.12}  &  8.35  &   \textbf{4.94}  \\
    JFA~\cite{xu2017joint} & 5.32  & 9.11  & 6.06 \\
    DAN~\cite{Kowalski2017DAN}   & 4.42  & 7.57  & 5.03 \\
    TR-DRN~\cite{lv2017deep} & 4.36  & \textbf{7.56}  & 4.99 \\
    DeFA~\cite{liu2017dense}  & 5.37  & 9.38  & 6.10 \\
    PIFA-S~\cite{jourabloo2017pose} & 5.43  & 9.88  & 6.30 \\
    RDR~\cite{xiao2017recurrent} & 5.03  & 8.95  & 5.80 \\
    \midrule
    ECT &  4.66  &  7.96 & 5.31 \\
    \bottomrule
    \end{tabular}%
  \end{center}
  \label{tab:comparison}%
\end{table}%

\begin{figure*}[t!]
    \centering
    \begin{subfigure}[h]{0.235\textwidth}
        \centering
        \includegraphics[width=1.1\textwidth]{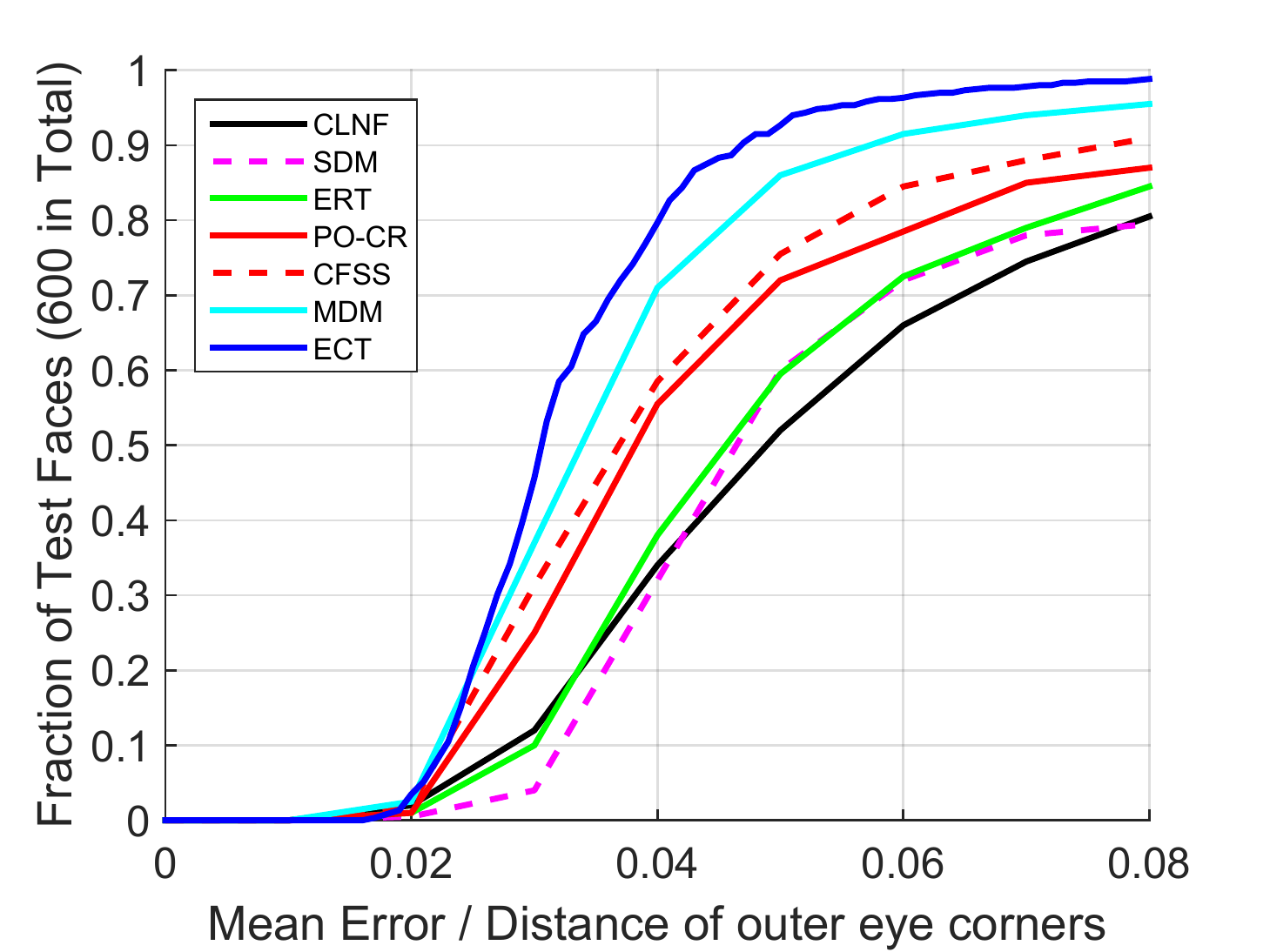}
        \caption{The 300W test set, 51 points}
        \label{fig:cprTest51}
    \end{subfigure}
    ~
    \begin{subfigure}[h]{0.235\textwidth}
        \centering
        \includegraphics[width=1.1\textwidth]{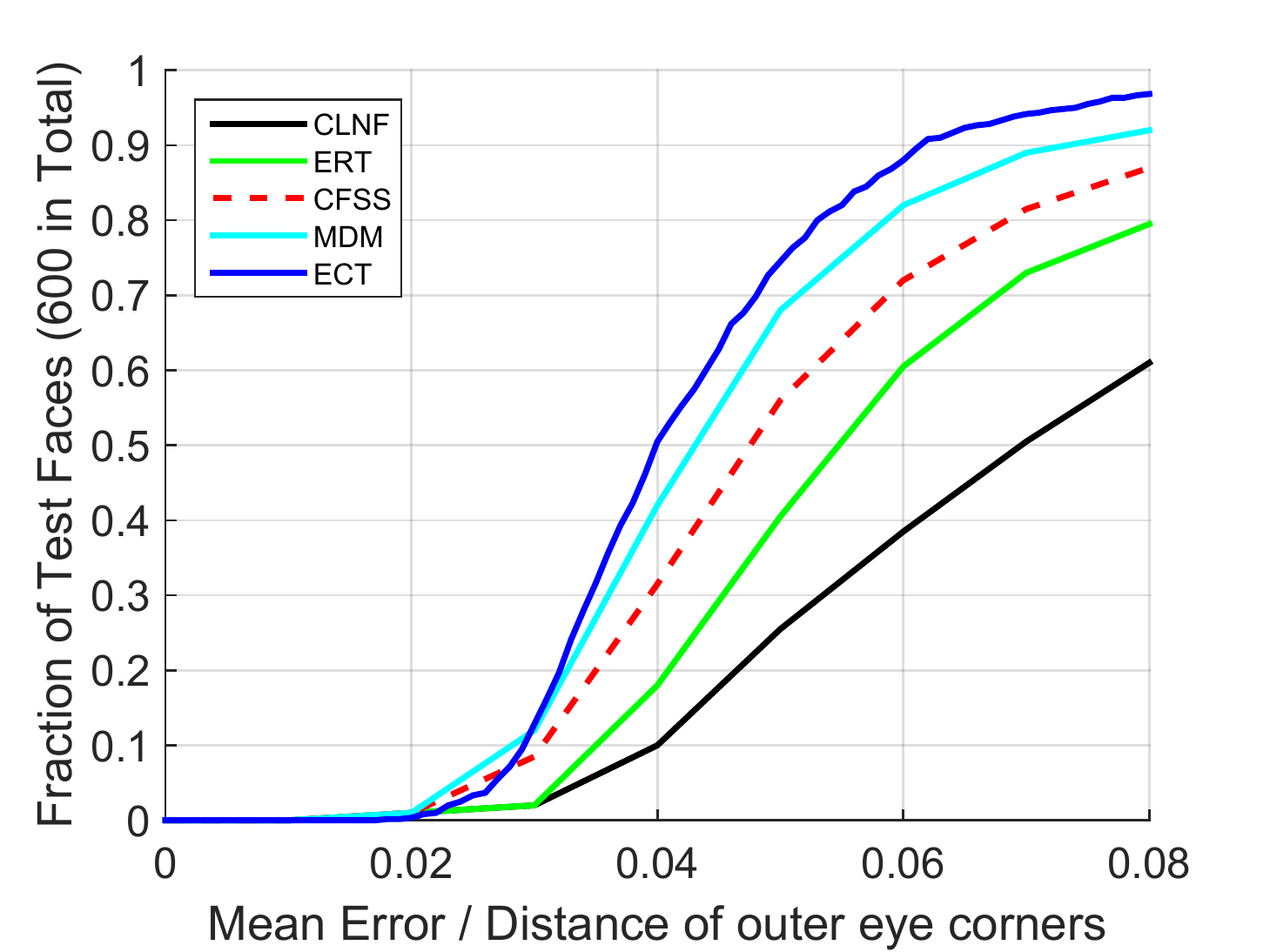}
        \caption{The 300W test set, 68 points}
        \label{fig:cprTest68}
    \end{subfigure}
    ~
    \begin{subfigure}[t!]{0.235\textwidth}
        \centering
        \includegraphics[width=1.1\textwidth]{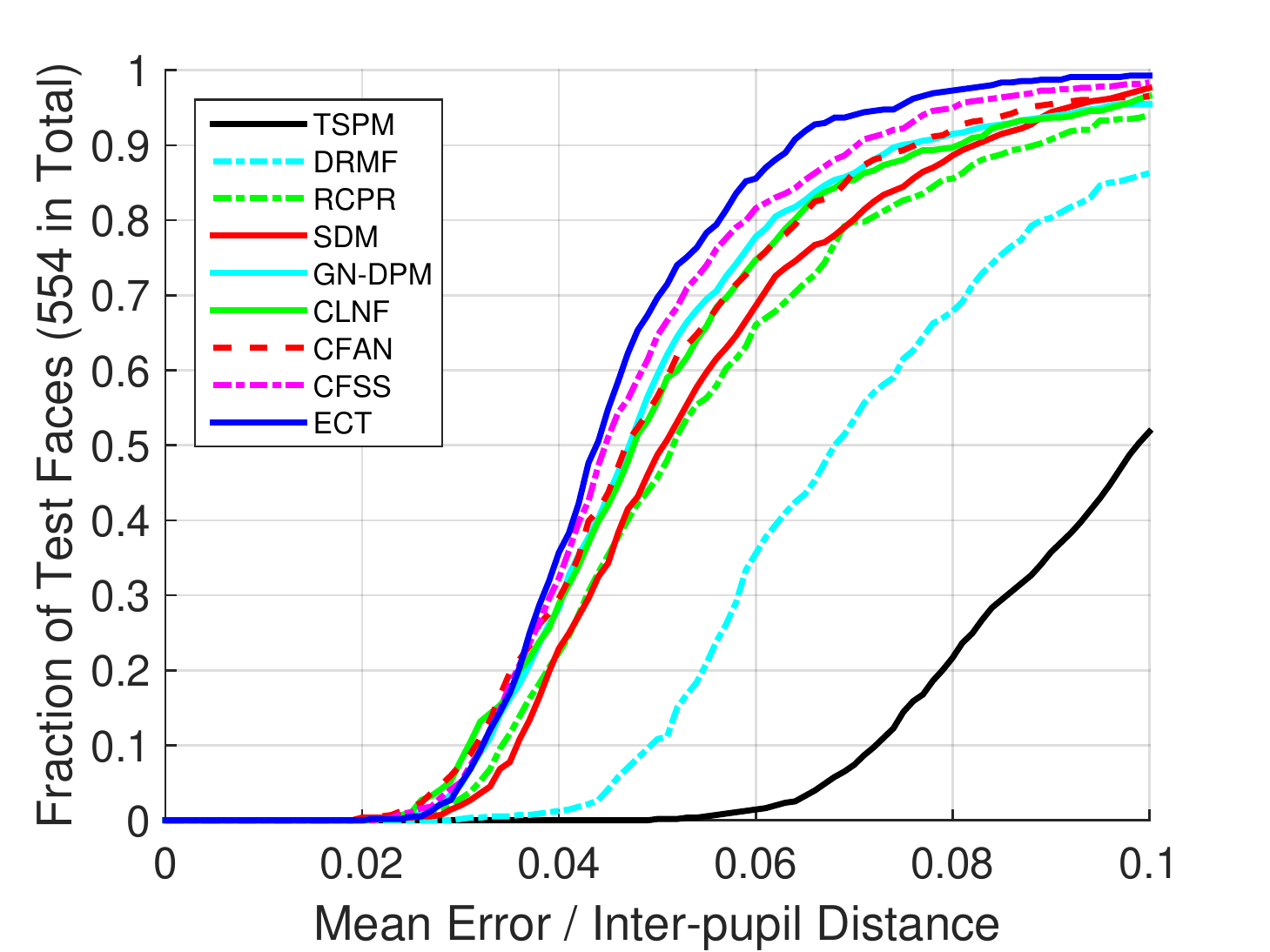}
        \caption{The common subset}
    \end{subfigure}%
    ~
    \begin{subfigure}[t!]{0.235\textwidth}
        \centering
        \includegraphics[width=1.1\textwidth]{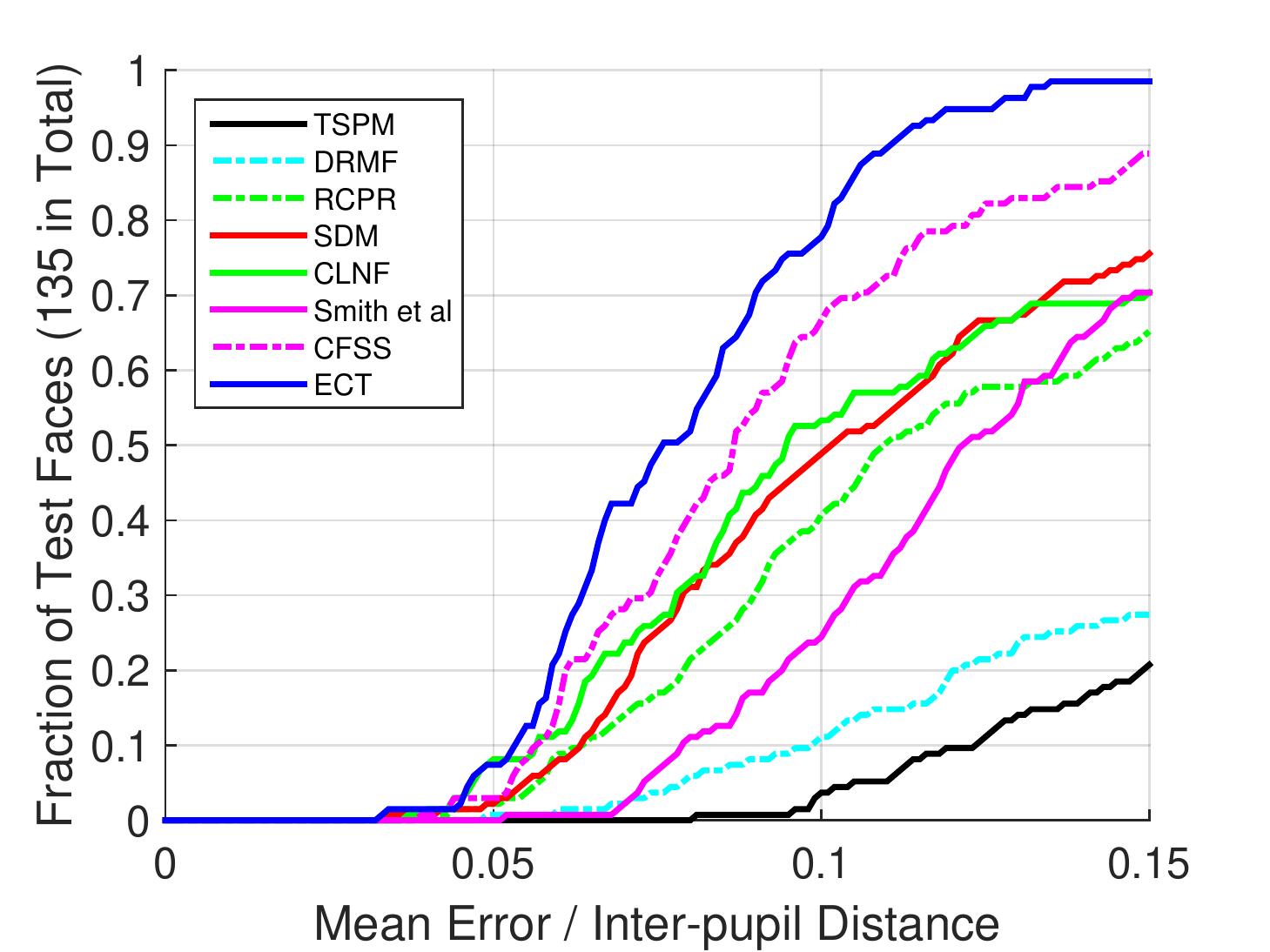}
        \caption{The challenging subset}
    \end{subfigure}
    \caption{Comparison of cumulative errors distribution (CED) curves on 300W. (a)(b) show CED curves of existing methods on the 300W test set for 51 points and 68 points. (c)(d) show CED curves of existing methods on the common subset and the challenging subset of 300W.%The proposed method outperforms state-of-the-art methods.
    %It should be noted that TCDCN uses additional training data besides the 300-W dataset.
        }
  \label{fig:comparisonStAr}
\end{figure*}

\subsection{Comparison with Existing Methods}

We compare our approach with existing methods including CLNF~\cite{baltrusaitis2013constrained}, CDM~\cite{yu2013pose}, SDM~\cite{xiong2013supervised}, LBF~\cite{ren2014face}, RCPR~\cite{burgos2013robust}, TCDCN~\cite{zhang2016learning}, DDN~\cite{yu2016deep}, MDM~\cite{trigeorgis2016mnemonic}, RAR~\cite{xiao2016robust}, PIFA-S~\cite{jourabloo2017pose}, RDR~\cite{xiao2017recurrent}, \etc.
Among these algorithms, CLNF and CDM are two part-based methods built upon the revised part models and parametric shape models.
SDM and LBF are two representative cascaded regression methods.
RCPR is a regression-based method aimed at handling occlusions.
TCDCN is a deep learning based method using multi-task learning.
DDN is a cascaded method incorporating structural constraints within the CNN framework.
MDM and RAR are two state-of-the-art methods using recurrent neural networks to refine the landmark prediction.
PIFA-S and RDR are two pose-invariant methods utilizing the 3D face model.

\subsubsection{Evaluation on 300W}

The evaluation on 300W consists of two parts.
The first part is conducted on the 300W test set provided officially by the 300W competition~\cite{sagonas2013300}.
The second part of the evaluation is performed on the fullset of 300W which is widely used in the literature.

For the 300W test set, the error is normalized by the distance of outer corners of the eyes to maintain consistency with the 300W competition.
The evaluation metrics used are the AUC and failure rate as in~\cite{trigeorgis2016mnemonic}.
The AUC stands for the area-under-the-curve of CED curves,
and the failure rate is calculated with the threshold set to 0.08 for the normalized point-to-point error.
%More details about the evaluation metric can refer to the 300W competition.
The comparison results with different methods for both 51 points and 68 points are reported in Table~\ref{tab:compr300WTest}.
Experimental results show that our method is narrowly beaten by DAN~\cite{Kowalski2017DAN} and outperforms other state-of-the-art methods including PO-CR~\cite{tzimiropoulos2015project}, CFSS~\cite{zhu2015face} and MDM~\cite{trigeorgis2016mnemonic} especially on failure rate.

For the fullset of 300W, we compare the localization results on the common subset, the challenging subset and the fullset.
For fair comparison, the localization error is normalized by the inter-pupil distance, which is consistent with previous works~\cite{ren2014face,zhu2015face}.
Since the pupil landmark positions are not available in the 300W dataset, they are instead estimated by averaging the coordinates of the landmarks around the eyes~\cite{ren2014face,zhu2015face}.
The comparison with various state-of-the-art methods are reported in Table~\ref{tab:comparison}.
It should be noted that the performance on the fullset of 300W is nearly saturated for the end-to-end methods proposed recently.
Our method achieves a comparable result in comparison with the most recent state-of-the-art methods RAR~\cite{xiao2016robust}, DAN~\cite{Kowalski2017DAN} and TR-DRN~\cite{lv2017deep}, and shows superior performance to others.

\begin{figure*}[t]
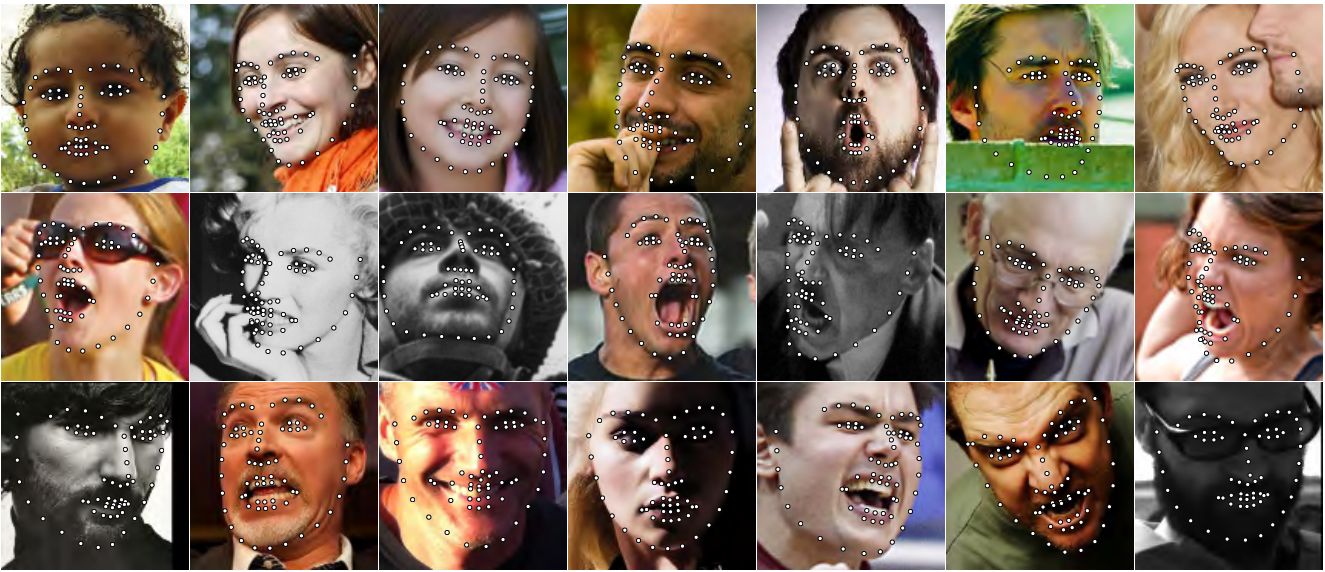

	\centering
	\foreach \idx in {1,2,3,4,5,6,7} {
		\begin{subfigure}[h]{0.125\textwidth}
			\centering
			\includegraphics[width=1.1\textwidth]{Img/demo/300w/300w_\idx.pdf}
		\end{subfigure}
	}
	
	\vspace{0.2mm}
	\foreach \idx in {8,9,10,11,12,13,14} {
		\begin{subfigure}[h]{0.125\textwidth}
			\centering
			\includegraphics[width=1.1\textwidth]{Img/demo/300w/300w_\idx.pdf}
		\end{subfigure}
	}
	
	\vspace{0.2mm}
	\foreach \idx in {15,16,17,18,19,20,21} {
		\begin{subfigure}[h]{0.125\textwidth}
			\centering
			\includegraphics[width=1.1\textwidth]{Img/demo/300w/300w_\idx.pdf}
		\end{subfigure}
	}
	\caption{Example results of the proposed method on the 300W dataset. The images are from the common subset (the top row), the challenging subset (the middle row), and the 300W testset (the bottom row), respectively. }
	\label{fig:300wDemo}
\end{figure*}

Cumulative error distribution (CED) curves of different methods on 300W are also plotted in Fig.~\ref{fig:comparisonStAr}.
As can be seen, the performance of ECT is superior to other approches particularly on the 300W test set and the challenging subset,
which means ECT is robust to various challenging conditions such as exaggerated expressions or occlusions.
Fig.~\ref{fig:300wDemo} shows example results of the proposed method on 300W.

\begin{table}[htbp]
	\color{black}
  \centering
    \caption{Comparison of NME (\%) with state-of-the-art methods on AFLW. Results of other methods are obtained from literature~\cite{zhu2016unconstrained,jourabloo2016large,bulat2016convolutional}.}
    \begin{tabular}{lcc}
    \toprule
          & \multicolumn{2}{c}{AFLW-PIFA} \\
\cmidrule{2-3}    Method & 21 points (vis.) & 34 points (vis.) \\
    \midrule
    CDM~\cite{yu2013pose}   & 8.59  & - \\
    RCPR~\cite{burgos2013robust}  & 7.15  & 6.26 \\
    CFSS~\cite{zhu2015face}  & 6.75  & - \\
    ERT~\cite{kazemi2014one}   & 7.03  & - \\
    SDM~\cite{xiong2013supervised}   & 6.96  & - \\
    LBF~\cite{ren2014face}   & 7.06  & - \\
    PIFA~\cite{jourabloo2015pose}  & 6.52  & 8.04 \\
    PAWF~\cite{jourabloo2016large}  & -     & 4.72 \\
    CCL~\cite{zhu2016unconstrained}   & 5.81  & - \\
    CALE~\cite{bulat2016convolutional}  & \textbf{2.63}  & \textbf{2.96} \\
    KEPLER~\cite{kumar2017kepler} & 2.98   & - \\
    PIFA-S~\cite{jourabloo2017pose} & -     & 4.45 \\
	DeFA~\cite{liu2017dense}  &  -     &   3.86 \\
    \midrule
    ECT   & 3.21 & 3.36 \\
    \bottomrule
    \end{tabular}%
  \label{tab:comparisonAFLW}%
\end{table}%

\subsubsection{Evaluation on AFLW}

\begin{figure*}[t]
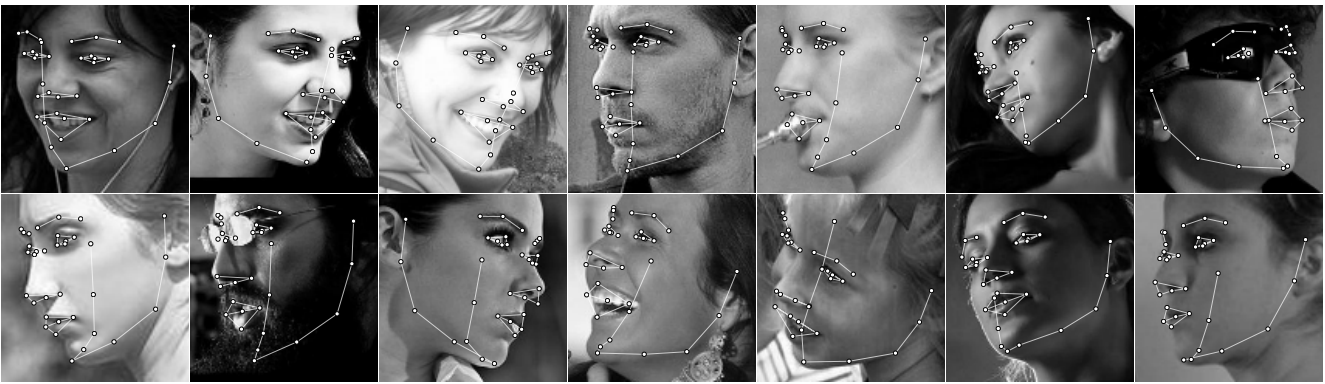

	\centering
	\foreach \idx in {1,2,3,4,5,6,7} {
		\begin{subfigure}[h]{0.125\textwidth}
			\centering
			\includegraphics[width=1.1\textwidth]{Img/demo/aflw/pifa_\idx.pdf}
		\end{subfigure}
	}
	
	\vspace{0.2mm}
	\foreach \idx in {8,9,10,11,12,13,14} {
		\begin{subfigure}[h]{0.125\textwidth}
			\centering
			\includegraphics[width=1.1\textwidth]{Img/demo/aflw/pifa_\idx.pdf}
		\end{subfigure}
	}
	\caption{Example results of the proposed method on the AFLW-PIFA dataset. Note that all landmarks are localized from a 3D perspective.}
	\label{fig:alfwDemo}
\end{figure*}

We evaluate our method on AFLW-PIFA~\cite{jourabloo2015pose} to demonstrate its effectiveness on face images with challenging appearance and head pose variations.
During testing, only the visible landmarks are involved in the evaluation.
To be consistent with previous works~\cite{jourabloo2015pose,zhu2016unconstrained}, the normalized distance is the square root of the bounding box size, calculated as $\sqrt{w_{bbox}*h_{bbox}}$.
The comparison with existing methods for both 21 and 34 points are shown in Table~\ref{tab:comparisonAFLW}.
It can be observed that ECT achieves results comparable to the latest state-of-the-art methods CALE~\cite{bulat2016convolutional}  and KEPLER~\cite{kumar2017kepler}, and outperforms pose-invariant approaches CDM~\cite{yu2013pose}, CCL~\cite{zhu2016unconstrained}, 3D approaches PIFA~\cite{jourabloo2015pose}, PAWF~\cite{jourabloo2016large} and PIFA-S~\cite{jourabloo2017pose} by a large margin.
It should be noted that CALE develops a much deeper neural network with the model size an order of magnitude larger than ours.
For comparison on landmark detection error across poses, three subsets are divided according to their absolute yaw angles: [0\degree, 30\degree], [30\degree, 60\degree], and [60\degree, 90\degree] with each subset containing 433 samples.
Comparison of landmark detection error of 21 points visible on AFLW across poses is reported in Table~\ref{tab:comparisonAFLWangle}.
Note that the results of RCPR, ESR, and SDM are derived from~\cite{zhu2016face} and these algorithms have been adapted to large poses by retraining them on 300W-LP~\cite{zhu2016face}.
As shown in Table~\ref{tab:comparisonAFLWangle}, the proposed method outperforms other algorithms consistently across poses.
Fig.~\ref{fig:alfwDemo} shows example results of the proposed method on the AFLW-PIFA dataset.

\begin{table}[t]
  \centering
  \caption{Comparison of NME (\%) on AFLW across poses. Results of other methods are obtained from~\cite{zhu2016face,ranjan2016hyperface,Bhagavatula2017Faster,xiao2017recurrent}.}
    \begin{tabular}{l|ccc|cc}
    \toprule
    %& \multicolumn{5}{c}{AFLW (21 pts)} \\
    %\cmidrule{2-6}
    Method & [0\degree, 30\degree]&[30\degree, 60\degree]&[60\degree, 90\degree]& mean  & std \\
    \midrule
    CDM~\cite{yu2013pose}   & 8.15  & 13.02 & 16.17 & 12.44 & 4.04 \\
    %RCPR~\cite{burgos2013robust}  & 6.16  & 18.67 & 34.82 & 19.88 & 14.36 \\
    RCPR~\cite{burgos2013robust} & 5.43  & 6.58  & 11.53 & 7.85  & 3.24 \\
    ESR~\cite{cao2014face} & 5.66  & 7.12  & 11.94 & 8.24  & 3.29 \\
    SDM~\cite{xiong2013supervised} & 4.75  & 5.55  & 9.34  & 6.55  & 2.45 \\
    3DDFA~\cite{zhu2016face} & 4.75  & 4.83  & 6.38  & 5.32  & 0.92 \\
    HyerFace~\cite{ranjan2016hyperface} & 3.93  & 4.14  & 4.71  & 4.26  & \textbf{0.41} \\
    3DSTN~\cite{Bhagavatula2017Faster} & 3.55  & 3.92  & 5.21  & 4.23  & 0.87 \\
    RDR~\cite{xiao2017recurrent} & 3.63  & 4.29  & 5.31  & 4.41  & - \\
    \midrule
    ECT   & \textbf{2.94}  & \textbf{2.84}  & \textbf{3.85}  & \textbf{3.21} & 0.56 \\
    \bottomrule
    \end{tabular}%
  \label{tab:comparisonAFLWangle}%
\end{table}%

\begin{table}[htbp]
	\centering
	\caption{Comparison of NME (\%) and MAPE (pixels) with state-of-the-art methods on AFW. Results of other methods are obtained from \cite{zhu2016unconstrained,jourabloo2016large} and respective papers.}
	\begin{tabular}{lcc}
		\toprule
		% & \multicolumn{2}{c}{AFW} \\
		% \cmidrule{2-3}    Method & NME (\%)   & MAPE (pixels) \\
		Method & NME (\%)   & MAPE (pixels) \\
		\midrule
	    TSPM~\cite{zhu2012face}  & -     & 11.09 \\
		CDM~\cite{yu2013pose}   & 5.70   & 9.13 \\
		RCPR~\cite{burgos2013robust}  & 3.87  & - \\
		CFSS~\cite{zhu2015face}  & 3.43  & - \\
		ERT~\cite{kazemi2014one}   & 3.25  & - \\
		SDM~\cite{xiong2013supervised}   & 3.88  & - \\
		LBF~\cite{ren2014face}   & 3.39  & - \\
		PIFA~\cite{jourabloo2015pose}  & -     & 8.61 \\
		PAWF~\cite{jourabloo2016large}  & -     & 7.43 \\
		CCL~\cite{zhu2016unconstrained}   & \textbf{2.45} & - \\
		KEPLER~\cite{kumar2017kepler} & 3.01  & - \\
		PIFA-S~\cite{jourabloo2017pose} & -     & 6.27 \\
		\midrule
		ECT   & 2.62  & \textbf{5.90} \\
		\bottomrule
	\end{tabular}%
	\label{tab:comparisonAFW}%
\end{table}%

\subsubsection{Evaluation on AFW}
\begin{figure*}[t]
	\centering
	\foreach \idx in {1,2,3,4,5,6,7} {
		\begin{subfigure}[h]{0.125\textwidth}
			\centering
			\includegraphics[width=1.1\textwidth]{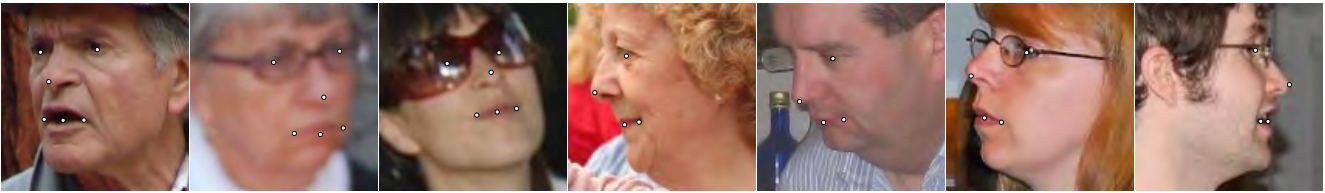}
		\end{subfigure}
	}
	\caption{Example results of the proposed method on the AFW dataset.}
	\label{fig:afwDemo}
\end{figure*}

We further test our method on AFW using the model trained on AFLW-PIFA.
Following the setting of previous works~\cite{yu2013pose,zhu2016unconstrained,jourabloo2016large}, we pick out 6 visible landmarks for evaluation on AFW.
We report both the Normalized Mean Error (NME) and the Mean Average Pixel Error (MAPE) for comparison in Table~\ref{tab:comparisonAFW}.
For NME, the normalized distance is the square root of the bounding box size provided in the AFW dataset.
For MAPE, the average of landmark detection errors is calculated at the original image scale.
As shown in Table~\ref{tab:comparisonAFW}, the proposed method achieves at least a comparable, or a superior performance in comparison with other state-of-the-art methods including CCL~\cite{zhu2016unconstrained}, KEPLER~\cite{kumar2017kepler} and PIFA-S~\cite{jourabloo2017pose}.
Example results of the proposed method on AFW is shown in Fig.~\ref{fig:afwDemo}.

\subsubsection{Evaluation on COFW}

\begin{figure*}[t]
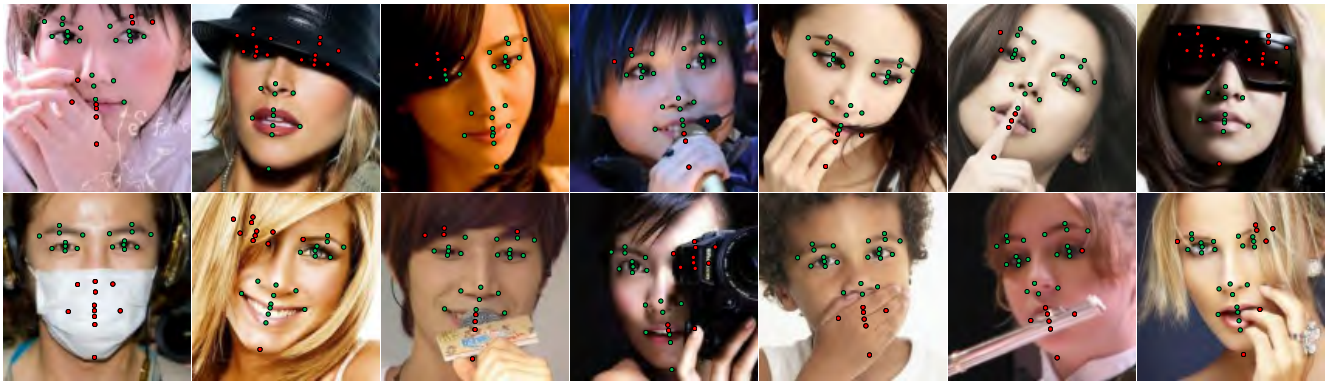

	\centering
	\foreach \idx in {1,2,3,4,5,6,7} {
		\begin{subfigure}[h]{0.125\textwidth}
			\centering
			\includegraphics[width=1.1\textwidth]{Img/demo/cofw/cofw_\idx.pdf}
		\end{subfigure}
	}
	
	\vspace{0.2mm}
	\foreach \idx in {8,9,10,11,12,13,14} {
		\begin{subfigure}[h]{0.125\textwidth}
			\centering
			\includegraphics[width=1.1\textwidth]{Img/demo/cofw/cofw_\idx.pdf}
		\end{subfigure}
	}
	\caption{Example results of the proposed method on the COFW dataset.  Green/red points indicate the visible/invisible landmarks predicted by the algorithm.}
	\label{fig:cofwDemo}
\end{figure*}

We conduct experiments on the COFW dataset to quantitatively demonstrate the effectiveness of the proposed method on face images with heavy occlusions.
%Experiments conducted on the COFW dataset is used to quantitatively demonstrate the effectiveness of the proposed method on face images with heavy occlusions.
507 face images from the test set of COFW are used for evaluation.
Table~\ref{tab:cprCOFW} shows the comparison of the landmark detection error, failure rate and occlusion prediction accuracy on COFW.
The mean error is normalized with respect to the inter-pupil distance and the failure rate is calculated with the threshold set to 10\% of the normalized mean error.
In our method, the occlusion status of the landmarks are simply predicted by thresholding the confidence (i.e. Eq.~\eqref{eq:sigmoidweight}) of the patch response maps in the final stage.
It can be seen that the proposed method achieves a nearly saturated landmark detection performance and a significantly higher occlusion prediction accuracy (recall of 63.4\% \vs 49.11\% at precision = 80\%) compared with the previous methods including Wu~\etal~\cite{wu2015robust}, RCPR~\cite{burgos2013robust}, RAR~\cite{xiao2016robust} and SimLPD~\cite{Wu2017Simultaneous}.
Example results of our method are depicted in Fig.~\ref{fig:cofwDemo}.

\begin{table}[t]
	\centering
	\caption{Comparison of facial landmark detection and occlusion prediction results on COFW. The results of TSPM, CDM, ESR, and SDM are obtained from~\cite{zhang2016learning}.}
	\begin{tabular}{lccc}
		\toprule
		%& \multicolumn{1}{c}{Landmark detection}  & \multicolumn{1}{c}{Occlusion prediction }\\
		%Method & \multicolumn{1}{c}{Normalized Mean Error \%} & \multicolumn{1}{c}{
		%Precision/Recall} & \multicolumn{1}{c}{Failure rate} \\
		Method & \multicolumn{1}{c}{NME (\%)} & \multicolumn{1}{c}{Failure Rate (\%)} & \multicolumn{1}{c}{Precision/Recall (\%)} \\
		\midrule
		Human~\cite{burgos2013robust}  & 5.60   & - & - \\
		\midrule
		TSPM~\cite{zhu2012face}  & 14.4  & -     & - \\
		CDM~\cite{yu2013pose}   & 13.67 & -     & - \\
		ESR~\cite{cao2014face}   & 11.2  & -     & - \\
		SDM~\cite{xiong2013supervised}   & 11.14 & -     & - \\
		RCPR~\cite{burgos2013robust}  & 8.5   & 20 & 80/40 \\
		OC~\cite{ghiasi2014occlusion}    & 7.46  & 13.24  & 80.8/37.0 \\
		RPP~\cite{yang2015robust}   & 7.52  & 16.20 & 78/40 \\
		Wu~\etal~\cite{wu2015robust} & \textbf{5.93}  & - & 80/49.11 \\
		TCDCN~\cite{zhang2016learning} & 8.05  & -     & - \\
		RAR~\cite{xiao2016robust}   & 6.03  & \textbf{4.14} & - \\
		SimLPD~\cite{Wu2017Simultaneous} & 6.40   & - & 80/44.43 \\
		\midrule
		ECT  & 5.98  & 4.54 & \textbf{80/63.4} \\
		\bottomrule
	\end{tabular}%
	\label{tab:cprCOFW}%
\end{table}%

\begin{table}[t]
	\centering
	\caption{Comparison of network complexity (number of parameters), runtime (time spent on GPU and CPU) and speed (faces per second) with other deep learning based methods. Numbers are obtained from respective papers or evaluated based on the released codes.}
	\begin{tabular}{l|cccc}
		\toprule
		&       & \multicolumn{2}{c}{Runtime} &  \\
		\cmidrule{3-4}    Method & \#Parameter & GPU   & CPU   & FPS \\
		\midrule
		CFAN~\cite{zhang2014coarse}  & $\sim$18M  & 0     & 23ms  & 43 \\
		TCDCN~\cite{zhang2016learning} & $\sim$0.1M & 0     & 18ms  & 56 \\
		3DDFA~\cite{zhu2016face} & -     & 23ms  & 52ms  & 13 \\
		DDN~\cite{yu2016deep}   & -     & -     & -     & 770 \\
		RAR~\cite{xiao2016robust}   & -     & -     & -     & 4 \\
		CALE~\cite{bulat2016convolutional}  &  $\sim$140M & -     & -     & 3 \\
		HyperFace~\cite{ranjan2016hyperface} & -     & -     & -     & 5 \\
		KEPLER~\cite{kumar2017kepler} & -     & -     & -     & 4 \\
		TR-DRN~\cite{lv2017deep} &  $\sim$99M  & -     & -     & 83 \\
		DAN~\cite{Kowalski2017DAN}   &  $\sim$22M  & -     & -     & 45 \\
		3DSTN~\cite{Bhagavatula2017Faster} & -     & -     & -     & 52 \\
		RDR~\cite{xiao2017recurrent} & -     & 31ms  & 142ms & 6 \\
		PIFA-S~\cite{jourabloo2017pose} & -     & -     & -     & 4.3 \\
		\midrule
		ECT   &  $\sim$9.5M & 30ms  & 53ms  & 12 \\
		\bottomrule
	\end{tabular}%
	\label{tab:Time}%
\end{table}%

\subsubsection{Time Complexity}
% In practical usage, 
Our method can run in realtime, with the speed of 12fps tested on an Intel Xeon 2.20GHz CPU and an NVIDIA TITAN X GPU.
Comparison of network complexity and runtime with other deep learning based methods is reported in Table~\ref{tab:Time}.
It can be seen that our method has a more moderate computation cost while achieving promising performances.
The most computationally expensive part of our method is generating the response maps.
It takes about 30ms for the FCN to process a 256x256 face image on GPU.
Using a shallower FCN could further reduce both the model size and runtime with only a slight drop in performance, as pointed out in the next subsection.
In our Python implementation of the weighed regularized mean shift, each iteration takes about 15ms to run on CPU.
Parallel processing of each landmark or conducting the weighted regularized mean shift on GPU could greatly reduce the runtime of the post-processing stage.
Tricks like using sparser candidate landmark sets or a precomputed grid for table lookup could further accelerate the fitting algorithm, as mentioned in~\cite{saragih2011deformable}.

\begin{figure*}[t]
    \begin{subfigure}[b]{0.325\textwidth}
        \centering
        \includegraphics[width=1.1\textwidth]{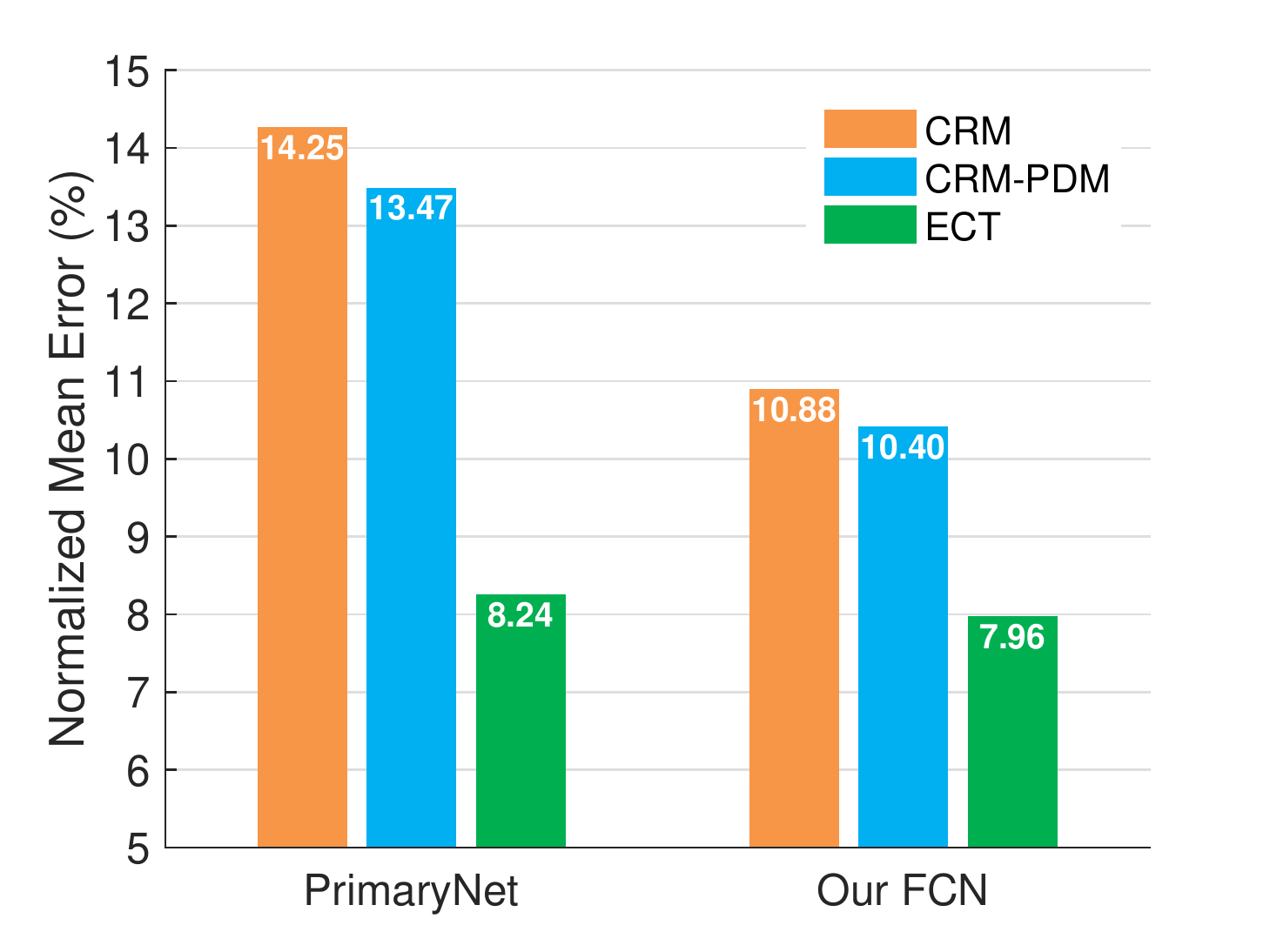}
        \caption{ }
        \label{fig:ablationFCN}
    \end{subfigure}
    \begin{subfigure}[b]{0.325\textwidth}
        \centering
        \includegraphics[width=1.1\textwidth]{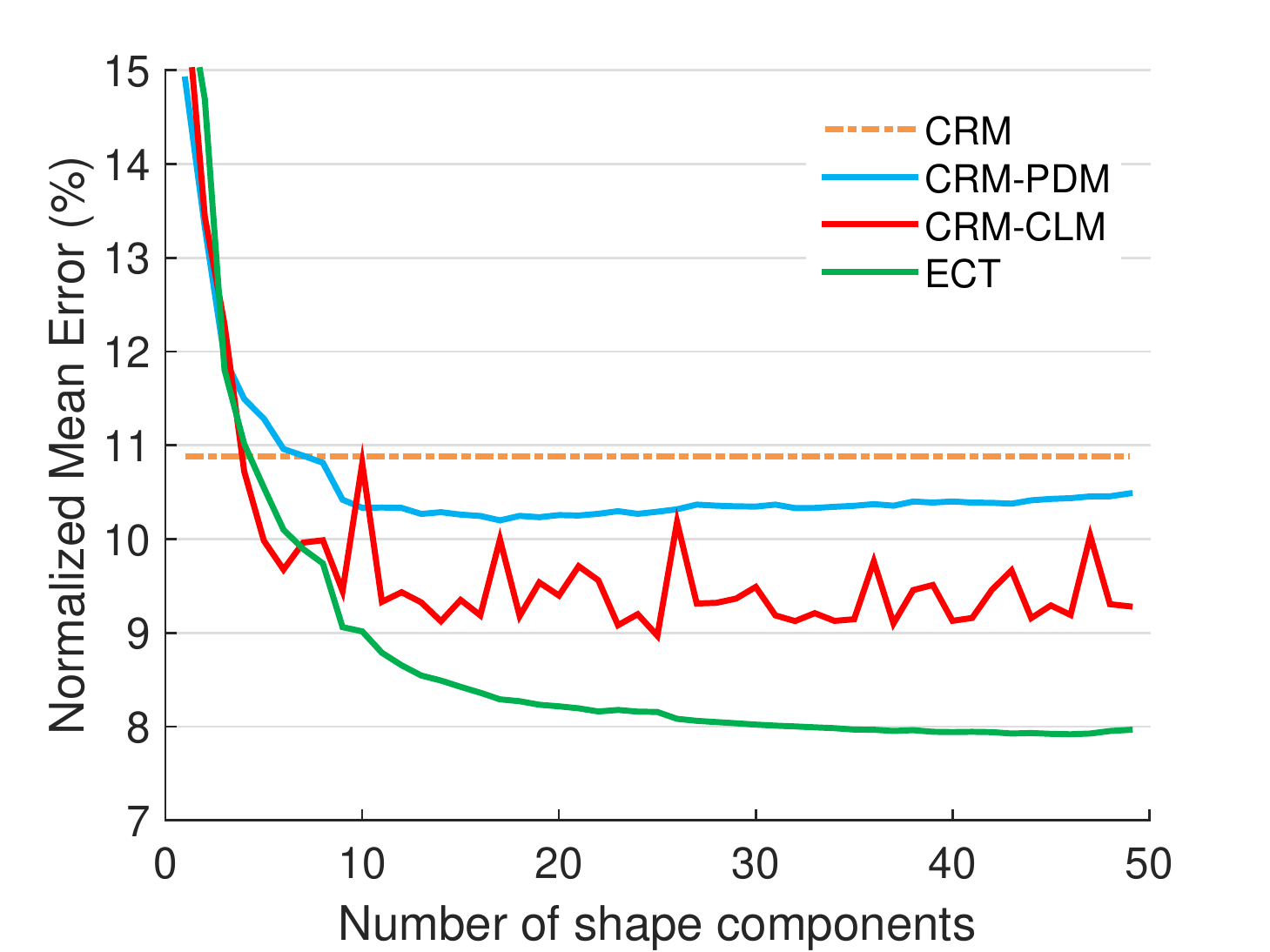}
        \caption{ }
        \label{fig:ablationPDM}
    \end{subfigure}
    \begin{subfigure}[b]{0.325\textwidth}
        \centering
        \includegraphics[width=1.1\textwidth]{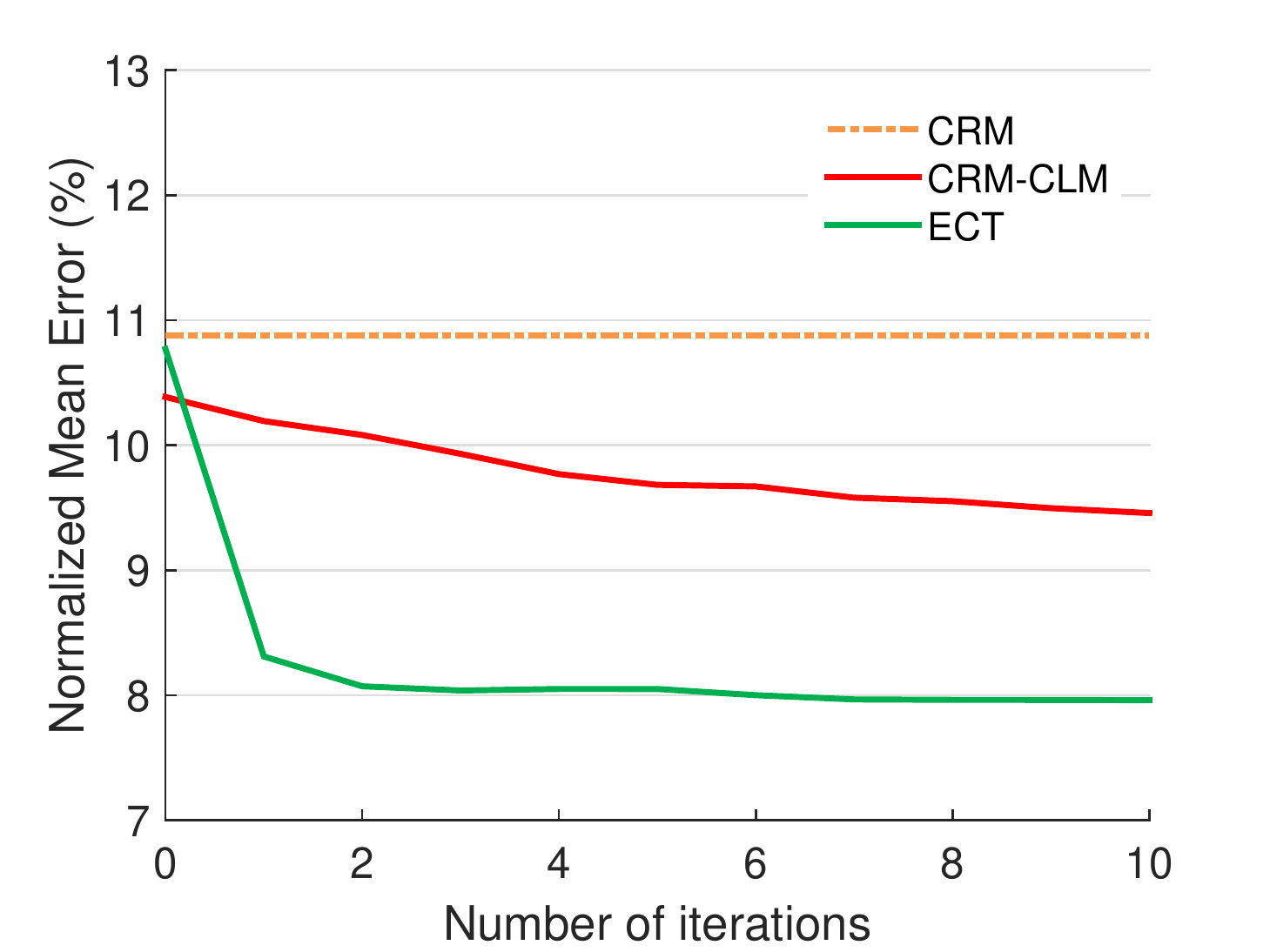}
        \caption{ }
        \label{fig:ablationIter}
    \end{subfigure}
    \caption{Component analysis of the proposed method on the challenging subset of 300W. (a) Performance of methods based on the proposed FCN and its subnetwork PrimaryNet. (b) Performance across the number of shape components. (c) Performance across iterations in the tuning step.}
\label{fig:Ablation}
\end{figure*}

\begin{table}[t]
  \caption{Comparison of NME (\%) to validate the key components of the proposed method.}
  \begin{center}
    \begin{tabular}{lccc}
    \toprule
    Experiment & \begin{tabular}[x]{@{}c@{}} Common\\ Subset\end{tabular}
    &  \begin{tabular}[x]{@{}c@{}} Challenging\\ Subset\end{tabular} & Fullset \\
    \midrule
    CRM(Baseline) & 6.61  & 10.88 & 7.44 \\
    CRM-PDM & 6.35  & 10.40 & 7.15 \\
    CRM-CLM & 5.28  & 9.49  & 6.10 \\
    CRM-CLM-C & 5.00  & 8.32  & 5.65 \\
    ECT &  4.66  &  7.96 & 5.31 \\
    \bottomrule
    \end{tabular}%
  \end{center}
  \label{tab:components}%
\end{table}%

\begin{figure*}[t]
   %\centering
        \centering
        \foreach \idx in {1,3,4,5} {
            \foreach \type in {img,heatmap,peak,pdm,init,ect} {
            \begin{subfigure}[h]{0.125\textwidth}
                \centering
                \includegraphics[width=1.1\textwidth]{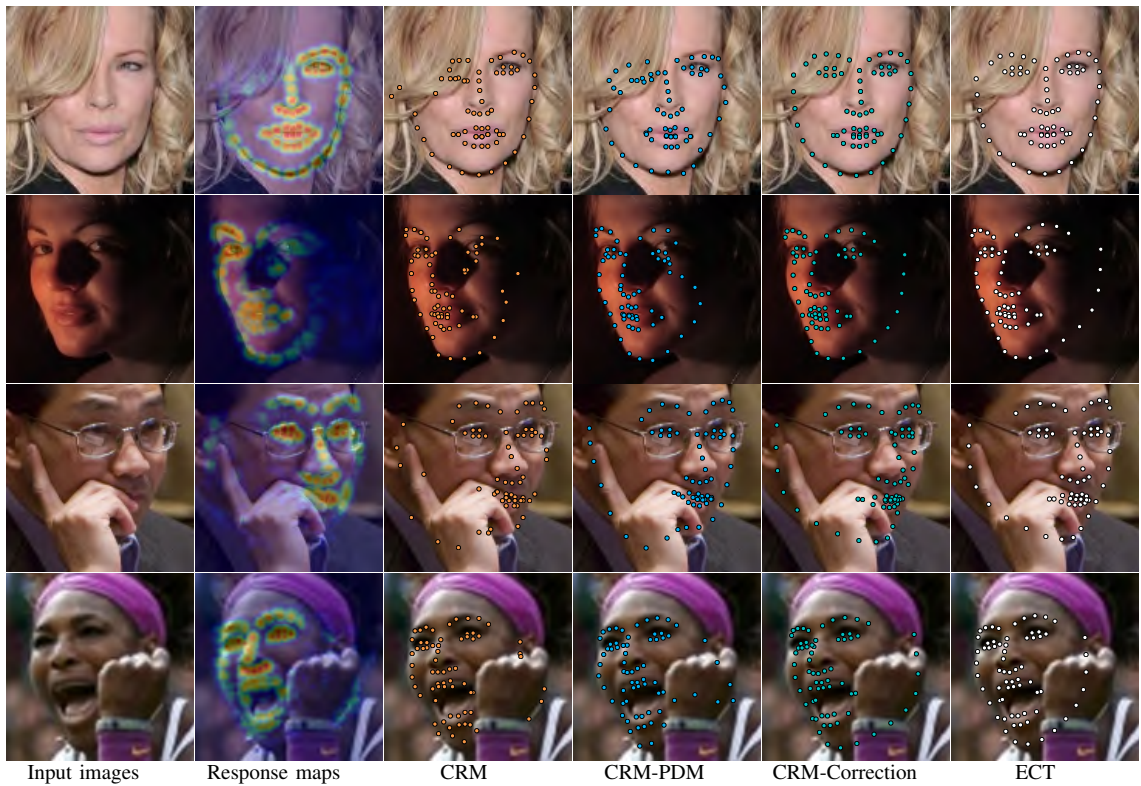}
            \end{subfigure}
            }
            \\
            \vspace{0.2mm}
        }
        \foreach \note in {Input images,Response maps,CRM,CRM-PDM,CRM-Correction,ECT} {
            \begin{subfigure}[h]{0.126\textwidth}
                \centering
                %\captionsetup{font=footnotesize}
                \caption*{\note}
            \end{subfigure}
        }
    \caption{Qualitative evaluation for ablation study of the proposed method. The images shown in Column 1 are input images. The response maps superimposed on the input images are shown in Column 2. The detection results of CRM and CRM-PDM are shown in Column 3 and 4, respectively. Column 5 shows the results obtained from the correction step of ECT. The results of the proposed ECT method are shown in the last column.}
\label{fig:AblationDemo}
\end{figure*}

\subsection{Ablation Study}

It is interesting to investigate the contribution of each module of the proposed ECT method.
In this subsection, we analyze the effectiveness of components of ECT using samples from the fullset of 300W.
The validation experiment results are shown in Table~\ref{tab:components}.
%Fig.~\ref{fig:AblationDemo} shows key components of the proposed method.
%The components analysis on the 300-W dataset is shown in Table~\ref{tab:components}.
CRM denotes the baseline method which simply locates the landmarks at the peak response positions in the Convolutional Response Maps.
The subsequent methods denoted in Table~\ref{tab:components} are the variant methods based on CRM, which will be introduced shortly.

\textbf{Validation of PDM.} CRM-PDM is a simple combination of CRM and PDM by projecting the peak response coordinates directly onto the PCA space of PDM. The improvement is consistent on both the common subset and the challenging subset in the comparison between CRM and CRM-PDM.
The results demonstrate the success of combining data-driven (CRM) and model-driven (PDM) methods.

\textbf{Validation of robust initialization.}
The correction step in our framework provides a reasonable initialization for subsequent procedures, which contributes to the robustness of the proposed method.
In Table~\ref{tab:components}, CRM-CLM uses the regularization results from CRM-PDM as the initial shape and refines the results using the original RLMS.
This approach could be regarded as a variant of CLM with its local experts extracted from our response maps.
CRM-CLM-C is similar to CRM-CLM but adopts the Correction step to initialize the landmark shape.
Comparison between CRM-CLM and CRM-CLM-C concludes that the proposed robust initialization could effectively improve the detection performance especially on the challenging subset.
%The improvements are 5.6\% and 12.3\% for the common subset and the challenging subset, respectively.

\textbf{Validation of the weighted regularized mean shift.}
%in Table~\ref{tab:components}
The aforementioned method CRM-CLM-C is equivalent to setting all weights as ones in Eq.~\eqref{eq:meanshift}.
A comparison between ECT and CRM-CLM-C shows that the tuning step based on weighted regularized mean shift is necessary to achieve a better performance.
%The mean errors can be decreased by 3.2\% and 4.3\% on the common subset and the challenging subset, respectively.
In addition, when comparing ECT and CRM-CLM in Table~\ref{tab:components}, we could observe the improvements of 12\% and 16\% on the common subset and the challenging subset, respectively.
Joint optimization of response maps and PDM in the weighted regularized mean shift framework leads to significant improvement on the challenging subset since the confidence of the response maps provides a more reasonable balance between the prior and the evidence.

Fig.~\ref{fig:AblationDemo} visualizes the response maps resulting from the baseline method and its variants for qualitative evaluation of key components of our method.
These results illustrate that ECT is capable of inferring the location of those blurred or invisible landmarks with the guidance of PDM prior and response maps.
%The improvement comes from not only the prior information for inferring the invisible landmarks, but also the weighted mean shift vectors.

\subsection{Component Analysis}

To gain insights of how each component contributes to the performance of the proposed method, we conduct experiments on the challenging subset (\ie IBUG) of 300W and report the results across different settings of each component.

\textbf{Performance across different FCN architectures.} Combinations with other FCN architectures rather than the proposed are also feasible in our framework.
The FCN proposed in Section~\ref{methodFCN} could be regarded as a light FCN and replaced with a shallower one.
To verify this, we remove the FusionNet from our FCN while keeping other sub-networks untact, and then finetune the truncated network with a small learning rate of $10^{-11}$.
The performance on the challenging subset based on these two architectures is reported in Fig.~\ref{fig:ablationFCN}.
The improvement over the baseline is more considerable though the new baseline is much worse.
The new FCN contains 30\% parameters less than the original one but still achieves a state-of-the-art performance on IBUG (with the mean error of 8.24), which demonstrates the potential for tailoring the proposed method to practical applications where the computational power is limited.

\textbf{Performance across the number of shape components.}
The structural information of facial landmarks is embedded in the PCA components of PDM.
Utilization of the global prior information is critical to part-based methods.
Fig.~\ref{fig:ablationPDM} shows the performance across different numbers of shape components.
It could be seen that there are obvious improvements over the baseline method CRM using within 10 shape components for CRM-PDM, CRM-CLM and ECT.
Compared with CRM-PDM and CRM-CLM, the proposed ECT could utilize more shape components for higher detection accuracy.
% the utilization of the PDM is limited to about 15 shape components for both CRM-PDM and CRM-CLM.

\textbf{Performance across iterations in the tuning step.}
The landmark detection errors at different iterations in the tuning step are reported in Fig.~\ref{fig:ablationIter}.
It is clear that the proposed method converges in two iterations, which is much faster than its counterpart CRM-CLM.
This could be attributed to the weighted strategy since the shape parameters update more efficiently in each iteration.

\textbf{Key Component Analysis.}
(\emph{i}) \emph{The convolutional response maps} regressed by FCN in a data-driven manner are highly discriminative and invariant to translation, scale and rotation.
As shown in Fig.~\ref{fig:landmarkweight} and~\ref{fig:AblationDemo}, the response maps are robust against large head poses, exaggerated expressions and scale variations in face images.
The invariance of the response maps makes great contributions to the robustness of our method to large pose and exaggerated expression.
Shape initialization from such desirable response maps could make full use of the holistic information so that our method is not prone to local minimums.
 %and compatible to different face detectors.
(\emph{ii}) \emph{The shape model PDM} encodes the prior structural information between facial landmarks.
It is used as a generative model to infer the occlusion part which is hard for the discriminative regressor FCN to deal with.
(\emph{iii}) \emph{The weighted regularized mean shift} incorporates the information from the regressor FCN and shape model PDM,
which could effectively balance the estimation effort of FCN and correction effort of PDM,
More intuitively, it could be observed from Fig.~\ref{fig:landmarkweight} and~\ref{fig:AblationDemo} that the landmark location lies close to the peak response point when the weight $w$ calculated from the convolutional response map is large, otherwise, it is found at the inference point spanned by PDM.

\section{Conclusion}
\label{Conclusion}

%Both model-driven and data-driven methods have been proposed for facial landmark detection in the literature. They both have some specific limitations.
In this work, we propose a three-step (Estimation-Correction-Tuning) framework, combining model-driven and data-driven methods, as a robust solution for facial landmark detection.
The proposed ECT method achieves superior, or at least comparable performance in comparison with state-of-the-art methods on challenging datasets including 300W, AFLW, AFW and COFW.
Experimental results demonstrate its effectiveness on face images with extreme appearance variations, large head poses and heavy occlusions.
The success of the method comes from holistically capturing the appearance information in a data-driven manner, explicitly utilizing the structural constraint in a model-driven manner and selectively balancing the efforts between the partial likelihood and global prior.
Basically, the proposed framework ECT manages to incorporate the discriminative regressor FCN with the generative model PDM, which is applicable to many similar problems in computer vision and pattern recognition problems. In future work, we will investigate ECT further in the context of general object alignment, human pose estimation, and image segmentation, \etc.

\appendices


\ifCLASSOPTIONcaptionsoff
  \newpage
\fi

\bibliographystyle{IEEEtran}
\bibliography{egbib}

\begin{thebibliography}{10}
\providecommand{\url}[1]{#1}
\csname url@samestyle\endcsname
\providecommand{\newblock}{\relax}
\providecommand{\bibinfo}[2]{#2}
\providecommand{\BIBentrySTDinterwordspacing}{\spaceskip=0pt\relax}
\providecommand{\BIBentryALTinterwordstretchfactor}{4}
\providecommand{\BIBentryALTinterwordspacing}{\spaceskip=\fontdimen2\font plus
\BIBentryALTinterwordstretchfactor\fontdimen3\font minus
  \fontdimen4\font\relax}
\providecommand{\BIBforeignlanguage}[2]{{%
\expandafter\ifx\csname l@#1\endcsname\relax
\typeout{** WARNING: IEEEtran.bst: No hyphenation pattern has been}%
\typeout{** loaded for the language `#1'. Using the pattern for}%
\typeout{** the default language instead.}%
\else
\language=\csname l@#1\endcsname
\fi
#2}}
\providecommand{\BIBdecl}{\relax}
\BIBdecl

\bibitem{cao2014face}
X.~Cao, Y.~Wei, F.~Wen, and J.~Sun, ``Face alignment by explicit shape
  regression,'' \emph{International Journal of Computer Vision}, vol. 107,
  no.~2, pp. 177--190, 2014.

\bibitem{xiong2013supervised}
X.~Xiong and F.~De~la Torre, ``Supervised descent method and its applications
  to face alignment,'' in \emph{Proceedings of the IEEE Conference on Computer
  Vision and Pattern Recognition}, 2013, pp. 532--539.

\bibitem{ren2014face}
S.~Ren, X.~Cao, Y.~Wei, and J.~Sun, ``Face alignment at 3000 fps via regressing
  local binary features,'' in \emph{Proceedings of the IEEE Conference on
  Computer Vision and Pattern Recognition}, 2014, pp. 1685--1692.

\bibitem{zhang2014coarse}
J.~Zhang, S.~Shan, M.~Kan, and X.~Chen, ``Coarse-to-fine auto-encoder networks
  (cfan) for real-time face alignment,'' in \emph{European Conference on
  Computer Vision}, 2014, pp. 1--16.

\bibitem{zhu2015face}
S.~Zhu, C.~Li, C.~Change~Loy, and X.~Tang, ``Face alignment by coarse-to-fine
  shape searching,'' in \emph{Proceedings of the IEEE Conference on Computer
  Vision and Pattern Recognition}, 2015, pp. 4998--5006.

\bibitem{zhang2016learning}
Z.~Zhang, P.~Luo, C.~C. Loy, and X.~Tang, ``Learning deep representation for
  face alignment with auxiliary attributes,'' \emph{IEEE Transactions on
  Pattern Analysis and Machine Intelligence}, vol.~38, no.~5, pp. 918--930,
  2016.

\bibitem{cootes1992active}
T.~F. Cootes and C.~J. Taylor, ``Active shape models¡ª¡®smart snakes¡¯,'' in
  \emph{British Machine Vision Conference}, 1992, pp. 266--275.

\bibitem{cootes2001active}
T.~F. Cootes, G.~J. Edwards, and C.~J. Taylor, ``Active appearance models,''
  \emph{IEEE Transactions on Pattern Analysis and Machine Intelligence}, no.~6,
  pp. 681--685, 2001.

\bibitem{tzimiropoulos2013optimization}
G.~Tzimiropoulos and M.~Pantic, ``Optimization problems for fast aam fitting
  in-the-wild,'' in \emph{Proceedings of the IEEE International Conference on
  Computer Vision}, 2013, pp. 593--600.

\bibitem{cristinacce2006feature}
D.~Cristinacce and T.~F. Cootes, ``Feature detection and tracking with
  constrained local models.'' in \emph{British Machine Vision Conference},
  vol.~1, no.~2, 2006, p.~3.

\bibitem{saragih2011deformable}
J.~M. Saragih, S.~Lucey, and J.~F. Cohn, ``Deformable model fitting by
  regularized landmark mean-shift,'' \emph{International Journal of Computer
  Vision}, vol.~91, no.~2, pp. 200--215, 2011.

\bibitem{sun2013deep}
Y.~Sun, X.~Wang, and X.~Tang, ``Deep convolutional network cascade for facial
  point detection,'' in \emph{Proceedings of the IEEE Conference on Computer
  Vision and Pattern Recognition}, 2013, pp. 3476--3483.

\bibitem{bulat2016convolutional}
A.~Bulat and G.~Tzimiropoulos, ``Convolutional aggregation of local evidence
  for large pose face alignment,'' in \emph{British Machine Vision Conference},
  2016, pp. 1--12.

\bibitem{trigeorgis2016mnemonic}
G.~Trigeorgis, P.~Snape, M.~A. Nicolaou, E.~Antonakos, and S.~Zafeiriou,
  ``Mnemonic descent method: A recurrent process applied for end-to-end face
  alignment,'' in \emph{Proceedings of the IEEE Conference on Computer Vision
  and Pattern Recognition}, 2016, pp. 4177--4187.

\bibitem{li2017fast}
Q.~Li, Z.~Sun, and R.~He, ``Fast multi-view face alignment via multi-task
  auto-encoders,'' in \emph{International Joint Conference on Biometrics},
  2017, pp. 538--545.

\bibitem{xiao2016robust}
S.~Xiao, J.~Feng, J.~Xing, H.~Lai, S.~Yan, and A.~Kassim, ``Robust facial
  landmark detection via recurrent attentive-refinement networks,'' in
  \emph{European Conference on Computer Vision}.\hskip 1em plus 0.5em minus
  0.4em\relax Springer, 2016, pp. 57--72.

\bibitem{lai2016deep}
H.~Lai, S.~Xiao, Y.~Pan, Z.~Cui, J.~Feng, C.~Xu, J.~Yin, and S.~Yan, ``Deep
  recurrent regression for facial landmark detection,'' \emph{IEEE Transactions
  on Circuits and Systems for Video Technology}, 2016.

\bibitem{Bulat2017HowFar}
A.~Bulat and G.~Tzimiropoulos, ``How far are we from solving the 2d \& 3d face
  alignment problem? (and a dataset of 230,000 3d facial landmarks),'' in
  \emph{Proceedings of the IEEE International Conference on Computer Vision},
  Oct 2017, pp. 1021--1030.

\bibitem{lv2017deep}
J.~Lv, X.~Shao, J.~Xing, C.~Cheng, and X.~Zhou, ``A deep regression
  architecture with two-stage re-initialization for high performance facial
  landmark detection,'' in \emph{Proceedings of the IEEE Conference on Computer
  Vision and Pattern Recognition}, 2017, pp. 3691--3700.

\bibitem{valstar2010facial}
M.~Valstar, B.~Martinez, X.~Binefa, and M.~Pantic, ``Facial point detection
  using boosted regression and graph models,'' in \emph{Proceedings of the IEEE
  Conference on Computer Vision and Pattern Recognition}.\hskip 1em plus 0.5em
  minus 0.4em\relax IEEE, 2010, pp. 2729--2736.

\bibitem{zhu2012face}
X.~Zhu and D.~Ramanan, ``Face detection, pose estimation, and landmark
  localization in the wild,'' in \emph{Proceedings of the IEEE Conference on
  Computer Vision and Pattern Recognition}, 2012, pp. 2879--2886.

\bibitem{pfister2015flowing}
T.~Pfister, J.~Charles, and A.~Zisserman, ``Flowing convnets for human pose
  estimation in videos,'' in \emph{Proceedings of the IEEE International
  Conference on Computer Vision}, 2015, pp. 1913--1921.

\bibitem{sagonas2013300}
C.~Sagonas, G.~Tzimiropoulos, S.~Zafeiriou, and M.~Pantic, ``300 faces
  in-the-wild challenge: The first facial landmark localization challenge,'' in
  \emph{Proceedings of the IEEE International Conference on Computer Vision
  Workshops}, 2013, pp. 397--403.

\bibitem{kostinger2011annotated}
M.~K{\"o}stinger, P.~Wohlhart, P.~M. Roth, and H.~Bischof, ``Annotated facial
  landmarks in the wild: A large-scale, real-world database for facial landmark
  localization,'' in \emph{Proceedings of the IEEE International Conference on
  Computer Vision Workshops}.\hskip 1em plus 0.5em minus 0.4em\relax IEEE,
  2011, pp. 2144--2151.

\bibitem{burgos2013robust}
X.~P. Burgos-Artizzu, P.~Perona, and P.~Doll{\'a}r, ``Robust face landmark
  estimation under occlusion,'' in \emph{Proceedings of the IEEE International
  Conference on Computer Vision}, 2013, pp. 1513--1520.

\bibitem{tzimiropoulos2014gauss}
G.~Tzimiropoulos and M.~Pantic, ``Gauss-newton deformable part models for face
  alignment in-the-wild,'' in \emph{Proceedings of the IEEE Conference on
  Computer Vision and Pattern Recognition}, 2014, pp. 1851--1858.

\bibitem{yu2013pose}
X.~Yu, J.~Huang, S.~Zhang, W.~Yan, and D.~N. Metaxas, ``Pose-free facial
  landmark fitting via optimized part mixtures and cascaded deformable shape
  model,'' in \emph{Proceedings of the IEEE International Conference on
  Computer Vision}, 2013, pp. 1944--1951.

\bibitem{asthana2013robust}
A.~Asthana, S.~Zafeiriou, S.~Cheng, and M.~Pantic, ``Robust discriminative
  response map fitting with constrained local models,'' in \emph{Proceedings of
  the IEEE Conference on Computer Vision and Pattern Recognition}, 2013, pp.
  3444--3451.

\bibitem{baltrusaitis2013constrained}
T.~Baltrusaitis, P.~Robinson, and L.-P. Morency, ``Constrained local neural
  fields for robust facial landmark detection in the wild,'' in
  \emph{Proceedings of the IEEE International Conference on Computer Vision
  Workshops}, 2013, pp. 354--361.

\bibitem{zadeh2017convolutional}
A.~Zadeh, Y.~C. Lim, T.~Baltru{\v{s}}aitis, and L.-P. Morency, ``Convolutional
  experts constrained local model for 3d facial landmark detection,'' in
  \emph{Proceedings of the IEEE Conference on Computer Vision and Pattern
  Recognition Workshops}, 2017, pp. 2051--2059.

\bibitem{alabort2015unifying}
J.~Alabort-i Medina and S.~Zafeiriou, ``Unifying holistic and parts-based
  deformable model fitting,'' in \emph{Proceedings of the IEEE Conference on
  Computer Vision and Pattern Recognition}, 2015, pp. 3679--3688.

\bibitem{xu2017joint}
X.~Xu and I.~A. Kakadiaris, ``Joint head pose estimation and face alignment
  framework using global and local cnn features,'' in \emph{12th IEEE
  International Conference on Automatic Face Gesture Recognition}, vol.~2,
  2017, pp. 642--649.

\bibitem{tompson2014joint}
J.~J. Tompson, A.~Jain, Y.~LeCun, and C.~Bregler, ``Joint training of a
  convolutional network and a graphical model for human pose estimation,'' in
  \emph{Advances in neural information processing systems}, 2014, pp.
  1799--1807.

\bibitem{newell2016stacked}
A.~Newell, K.~Yang, and J.~Deng, ``Stacked hourglass networks for human pose
  estimation,'' in \emph{European Conference on Computer Vision}.\hskip 1em
  plus 0.5em minus 0.4em\relax Springer, 2016, pp. 483--499.

\bibitem{huang2015densebox}
L.~Huang, Y.~Yang, Y.~Deng, and Y.~Yu, ``Densebox: Unifying landmark
  localization with end to end object detection,'' \emph{arXiv preprint
  arXiv:1509.04874}, 2015.

\bibitem{yang2017stacked}
J.~Yang, Q.~Liu, and K.~Zhang, ``Stacked hourglass network for robust facial
  landmark localisation,'' in \emph{Proceedings of the IEEE Conference on
  Computer Vision and Pattern Recognition Workshops}.\hskip 1em plus 0.5em
  minus 0.4em\relax IEEE, 2017, pp. 2025--2033.

\bibitem{Kowalski2017DAN}
M.~Kowalski, J.~Naruniec, and T.~Trzcinski, ``Deep alignment network: A
  convolutional neural network for robust face alignment,'' in
  \emph{Proceedings of the IEEE Conference on Computer Vision and Pattern
  Recognition Workshops}, July 2017, pp. 2034--2043.

\bibitem{kumar2017kepler}
A.~Kumar, A.~Alavi, and R.~Chellappa, ``Kepler: Keypoint and pose estimation of
  unconstrained faces by learning efficient h-cnn regressors,'' in \emph{12th
  IEEE International Conference on Automatic Face Gesture Recognition}, 2017,
  pp. 258--265.

\bibitem{szegedy2015going}
C.~Szegedy, W.~Liu, Y.~Jia, P.~Sermanet, S.~Reed, D.~Anguelov, D.~Erhan,
  V.~Vanhoucke, and A.~Rabinovich, ``Going deeper with convolutions,'' in
  \emph{Proceedings of the IEEE Conference on Computer Vision and Pattern
  Recognition}, 2015, pp. 1--9.

\bibitem{YuKoltun2016}
F.~Yu and V.~Koltun, ``Multi-scale context aggregation by dilated
  convolutions,'' in \emph{Proceedings of the International Conference on
  Learning Representations}, 2016.

\bibitem{matthews2004active}
I.~Matthews and S.~Baker, ``Active appearance models revisited,''
  \emph{International Journal of Computer Vision}, vol.~60, no.~2, pp.
  135--164, 2004.

\bibitem{belhumeur2013localizing}
P.~N. Belhumeur, D.~W. Jacobs, D.~J. Kriegman, and N.~Kumar, ``Localizing parts
  of faces using a consensus of exemplars,'' \emph{IEEE Transactions on Pattern
  Analysis and Machine Intelligence}, vol.~35, no.~12, pp. 2930--2940, 2013.

\bibitem{le2012interactive}
V.~Le, J.~Brandt, Z.~Lin, L.~Bourdev, and T.~S. Huang, ``Interactive facial
  feature localization,'' in \emph{European Conference on Computer Vision},
  2012, pp. 679--692.

\bibitem{jourabloo2015pose}
A.~Jourabloo and X.~Liu, ``Pose-invariant 3d face alignment,'' in
  \emph{Proceedings of the IEEE International Conference on Computer Vision},
  2015, pp. 3694--3702.

\bibitem{jourabloo2016large}
------, ``Large-pose face alignment via cnn-based dense 3d model fitting,'' in
  \emph{Proceedings of the IEEE Conference on Computer Vision and Pattern
  Recognition}, 2016, pp. 4188--4196.

\bibitem{zhu2016unconstrained}
S.~Zhu, C.~Li, C.-C. Loy, and X.~Tang, ``Unconstrained face alignment via
  cascaded compositional learning,'' in \emph{Proceedings of the IEEE
  Conference on Computer Vision and Pattern Recognition}, 2016, pp. 3409--3417.

\bibitem{jia2014caffe}
Y.~Jia, E.~Shelhamer, J.~Donahue, S.~Karayev, J.~Long, R.~Girshick,
  S.~Guadarrama, and T.~Darrell, ``Caffe: Convolutional architecture for fast
  feature embedding,'' in \emph{Proceedings of the 22nd ACM international
  conference on Multimedia}.\hskip 1em plus 0.5em minus 0.4em\relax ACM, 2014,
  pp. 675--678.

\bibitem{alabort2014menpo}
J.~Alabort-i Medina, E.~Antonakos, J.~Booth, P.~Snape, and S.~Zafeiriou,
  ``Menpo: A comprehensive platform for parametric image alignment and visual
  deformable models,'' in \emph{Proceedings of the 22nd ACM international
  conference on Multimedia}.\hskip 1em plus 0.5em minus 0.4em\relax ACM, 2014,
  pp. 679--682.

\bibitem{kazemi2014one}
V.~Kazemi and J.~Sullivan, ``One millisecond face alignment with an ensemble of
  regression trees,'' in \emph{Proceedings of the IEEE Conference on Computer
  Vision and Pattern Recognition}, 2014, pp. 1867--1874.

\bibitem{tzimiropoulos2015project}
G.~Tzimiropoulos, ``Project-out cascaded regression with an application to face
  alignment,'' in \emph{Proceedings of the IEEE Conference on Computer Vision
  and Pattern Recognition}, 2015, pp. 3659--3667.

\bibitem{smith2014nonparametric}
B.~M. Smith, J.~Brandt, Z.~Lin, and L.~Zhang, ``Nonparametric context modeling
  of local appearance for pose-and expression-robust facial landmark
  localization,'' in \emph{Proceedings of the IEEE Conference on Computer
  Vision and Pattern Recognition}, 2014, pp. 1741--1748.

\bibitem{zhu2016face}
X.~Zhu, Z.~Lei, X.~Liu, H.~Shi, and S.~Z. Li, ``Face alignment across large
  poses: A 3d solution,'' in \emph{Proceedings of the IEEE Conference on
  Computer Vision and Pattern Recognition}, 2016, pp. 146--155.

\bibitem{yu2016deep}
X.~Yu, F.~Zhou, and M.~Chandraker, ``Deep deformation network for object
  landmark localization,'' in \emph{European Conference on Computer
  Vision}.\hskip 1em plus 0.5em minus 0.4em\relax Springer, 2016, pp. 52--70.

\bibitem{liu2017dense}
Y.~Liu, A.~Jourabloo, W.~Ren, and X.~Liu, ``Dense face alignment,'' in
  \emph{Proceedings of the IEEE International Conference on Computer Vision
  Workshops}, 2017.

\bibitem{jourabloo2017pose}
A.~Jourabloo, M.~Ye, X.~Liu, and L.~Ren, ``Pose-invariant face alignment with a
  single cnn,'' in \emph{Proceedings of the IEEE International Conference on
  Computer Vision}, Oct 2017, pp. 3219--3228.

\bibitem{xiao2017recurrent}
S.~Xiao, J.~Feng, L.~Liu, X.~Nie, W.~Wang, S.~Yan, and A.~Kassim, ``Recurrent
  3d-2d dual learning for large-pose facial landmark detection,'' in
  \emph{Proceedings of the IEEE International Conference on Computer Vision},
  2017, pp. 1642--1651.

\bibitem{ranjan2016hyperface}
R.~Ranjan, V.~M. Patel, and R.~Chellappa, ``Hyperface: A deep multi-task
  learning framework for face detection, landmark localization, pose
  estimation, and gender recognition,'' \emph{IEEE Transactions on Pattern
  Analysis and Machine Intelligence}, 2017.

\bibitem{Bhagavatula2017Faster}
C.~Bhagavatula, C.~Zhu, K.~Luu, and M.~Savvides, ``Faster than real-time facial
  alignment: A 3d spatial transformer network approach in unconstrained
  poses,'' in \emph{Proceedings of the IEEE International Conference on
  Computer Vision}, Oct 2017, pp. 4000--4009.

\bibitem{wu2015robust}
Y.~Wu and Q.~Ji, ``Robust facial landmark detection under significant head
  poses and occlusion,'' in \emph{Proceedings of the IEEE International
  Conference on Computer Vision}, 2015, pp. 3658--3666.

\bibitem{Wu2017Simultaneous}
Y.~Wu, C.~Gou, and Q.~Ji, ``Simultaneous facial landmark detection, pose and
  deformation estimation under facial occlusion,'' in \emph{Proceedings of the
  IEEE Conference on Computer Vision and Pattern Recognition}, July 2017, pp.
  5719--5728.

\bibitem{ghiasi2014occlusion}
G.~Ghiasi and C.~C. Fowlkes, ``Occlusion coherence: Localizing occluded faces
  with a hierarchical deformable part model,'' in \emph{Proceedings of the IEEE
  Conference on Computer Vision and Pattern Recognition}, 2014, pp. 2385--2392.

\bibitem{yang2015robust}
H.~Yang, X.~He, X.~Jia, and I.~Patras, ``Robust face alignment under occlusion
  via regional predictive power estimation,'' \emph{IEEE Transactions on Image
  Processing}, vol.~24, no.~8, pp. 2393--2403, 2015.

\end{thebibliography}


\end{document}